\documentclass{article}

\usepackage{microtype}
\usepackage{graphicx}
\usepackage{subfigure}
\usepackage{booktabs} % for professional tables

\usepackage{hyperref}

\usepackage[accepted]{icml2023}

\usepackage{amsmath}
\usepackage{amssymb}
\usepackage{mathtools}
\usepackage{amsthm}
\usepackage[utf8]{inputenc} % allow utf-8 input
\usepackage[T1]{fontenc}    % use 8-bit T1 fonts
\usepackage{hyperref}       % hyperlinks
\usepackage{url}            % simple URL typesetting
\usepackage{booktabs}       % professional-quality tables
\usepackage{amsfonts}       % blackboard math symbols
\usepackage{nicefrac}       % compact symbols for 1/2, etc.
\usepackage{microtype}      % microtypography
\usepackage{xcolor}         % colors
\usepackage{graphicx}
\usepackage{wrapfig}
\usepackage{lipsum}
\usepackage{multirow}
\usepackage{textcomp}
\usepackage{sidecap}
\usepackage[toc,page,header]{appendix}
\usepackage{minitoc}

\usepackage{comment}

\usepackage[capitalize,noabbrev]{cleveref}

\theoremstyle{plain}
\newtheorem{theorem}{Theorem}[section]

\theoremstyle{definition}

\theoremstyle{remark}

\usepackage{Definitions}

\newcommand{\hf}{{\hat f}}
\newcommand{\hFcal}{{\hat \Fcal}}

\newcommand{\by}{{\bar y}}
\newcommand{\bx}{{\bar x}}
\newcommand{\tx}{{\tilde x}}

\newcommand{\answerTODO}[1][]{\textcolor{red}{\bf [TODO]}}

\usepackage[textsize=tiny]{todonotes}

\icmltitlerunning{Discrete Key-Value Bottleneck}

\begin{document}

\twocolumn[
\icmltitle{Discrete Key-Value Bottleneck}




\begin{icmlauthorlist}
\icmlauthor{Frederik Tr{\"a}uble}{1}
\icmlauthor{Anirudh Goyal}{6}
\icmlauthor{Nasim Rahaman}{1,2}
\icmlauthor{Michael Mozer}{4}
\\
\icmlauthor{Kenji Kawaguchi}{5}
\icmlauthor{Yoshua Bengio}{2,3,7}
\icmlauthor{Bernhard Sch{\"o}lkopf}{1,7}
\end{icmlauthorlist}

\icmlaffiliation{1}{MPI for Intelligent Systems, Tübingen}
\icmlaffiliation{2}{Mila} 
\icmlaffiliation{3}{Université de Montréal}
\icmlaffiliation{4}{Google Research, Brain Team}
\icmlaffiliation{5}{National University of Singapore}
\icmlaffiliation{6}{Google DeepMind}
\icmlaffiliation{7}{CIFAR Fellow}

\icmlcorrespondingauthor{Frederik Tr{\"a}uble}{frederik.traeuble@tuebingen.mpg.de}

\icmlkeywords{Machine Learning, ICML}

\vskip 0.3in
]



\printAffiliationsAndNotice{} % otherwise use the standard text.

\begin{abstract}
Deep neural networks perform well on classification tasks where data streams are i.i.d.\ and labeled data is abundant.
Challenges emerge with non-stationary training data streams such as continual learning. One powerful approach that has addressed this challenge involves pre-training of large encoders on volumes of readily available data, followed by task-specific tuning. Given a new task, however, updating the weights of these encoders is challenging as a large number of weights needs to be fine-tuned, and as a result, they forget information about the previous tasks. In the present work, we propose a model architecture to address this issue, building upon a discrete bottleneck containing  pairs of separate and learnable key-value codes. Our paradigm will be to \textit{encode; process the representation via a discrete bottleneck; and decode}. Here, the input is fed to the pre-trained encoder, the output of the encoder is used to select the nearest keys, and the corresponding values are fed to the decoder to solve the current task. The model can only fetch and re-use a sparse number of these key-value pairs during inference, enabling \textit{localized and context-dependent model updates}. We theoretically investigate the ability of the discrete key-value bottleneck to minimize the effect of learning under distribution shifts and show that it reduces the complexity of the hypothesis class. We empirically verify the proposed method under challenging class-incremental learning scenarios and show that the proposed model --- without any task boundaries --- reduces catastrophic forgetting across a wide variety of pre-trained models, outperforming relevant baselines on this task. 

\end{abstract}

\section{Introduction}
Current neural networks achieve state-of-the-art results across a range of challenging problem domains. Most of these advances are limited to learning scenarios where the training data is sampled i.i.d.\ from some assumed joint distribution $P(X,Y)=P(Y|X)P(X)$ in the supervised case. In reality, learning scenarios are often far from i.i.d.\ and exhibit various distribution shifts; for instance during continual learning, %or more generally 
when the training distribution changes over time. \looseness=-1

Despite all successes, the training of deep neural networks on non-stationary training data streams remains challenging. For instance, training neural networks without additional modifications on some small dataset or input distribution may quickly lead to over-fitting and catastrophic forgetting of previous knowledge \cite{chen2018lifelong, thrun1995learning, van2019three}. One paradigm that has dominated in recent work is the use of models pre-trained on large amounts of data e.g.\,in a self-supervised manner \cite{chen2020big, azizi2021big, caron2021emerging, brown2020language}. These models are then re-purposed for smaller datasets via either (a) fine-tuning, which may involve changing large numbers of parameters, or (b) introducing a small number of new parameters to adapt the model \cite{houlsby2019parameter, zhou2021learning, wang2021learning}. %\newline 

\begin{figure*}
    \centering
    \includegraphics[width=\textwidth]{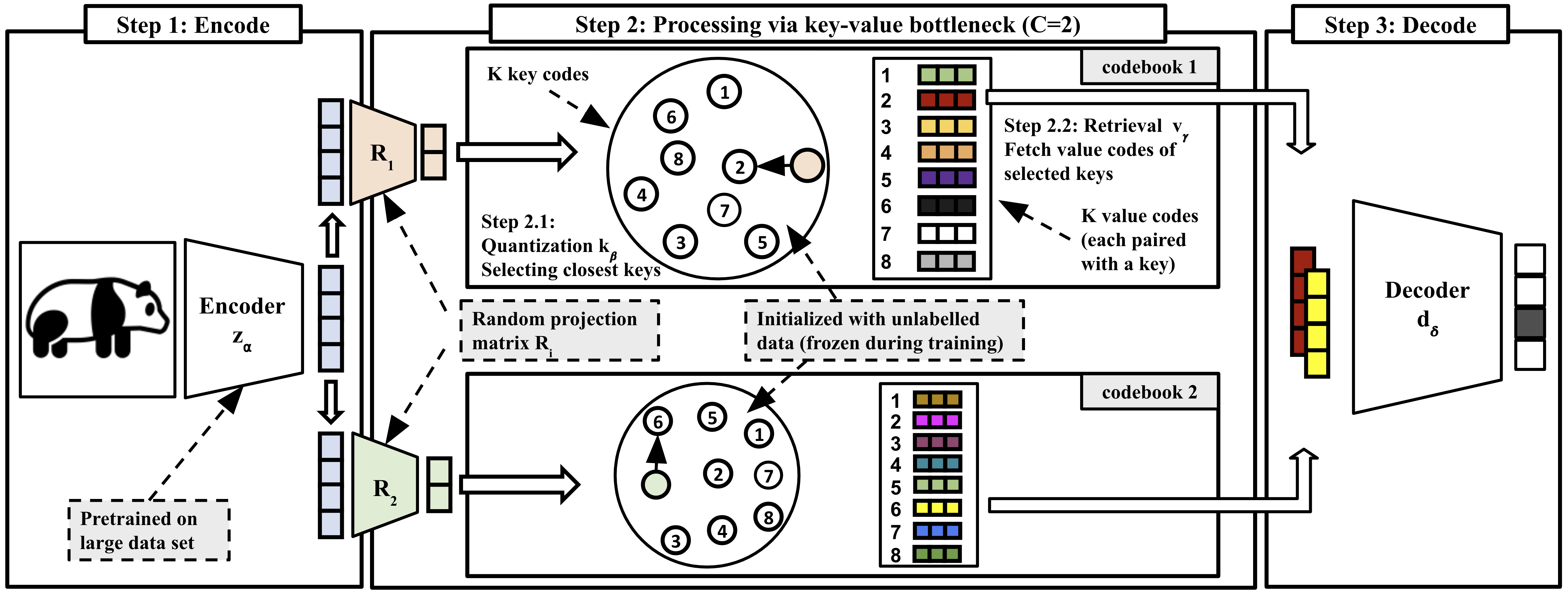}
    \caption{\textbf{Overview of discrete key-value bottleneck}. First, a pre-trained encoder maps the high-dimensional input to a lower-dimensional representation. Second, this representation is projected into $C$ lower-dimensional heads. $C=2$ is chosen for illustrative purposes. 
    Each head is processed as input by one out of $C$ separate key-value codebooks: In the codebook, the head is discretized by snapping to the closest key code, and the corresponding value code for that key is fetched. Third, the values from different key-value codebooks are combined and fed as input to the decoder for the final prediction. Keys and values can be of different dimensions in the proposed model.\looseness=-1}
    \label{fig:fig1}
\end{figure*}

Building upon this pre-training paradigm, we introduce an approach (see \Cref{fig:fig1} for an overview of the model architecture) to distil information into a discrete set of code pairs, each consisting of a coupled key and value code. In a nutshell, we follow a three-step process: \textit{encode, process via a discrete bottleneck, and decode}. %\newline 

In the first step, an input is fed to the encoder to obtain a continuous-valued representation that is projected into $C$ lower-dimensional heads. In the second step, these heads are used to search for the closest key within head-specific key-value codebooks, and \textit{corresponding} continuously learnable value codes taken from a discrete set are fetched. In the third step, these value codes are passed through a downstream decoder to produce the final output. By freezing all components except for the value codes, the model can continue to improve its performance \textit{locally} on input-dependent value codes without affecting the prediction and key-value pairs of prior training samples. %\newline 

As we will show theoretically and empirically, this architecture offers an appealing ability to improve generalization under input distribution shifts (also called \emph{covariate shifts}) 
without common vulnerabilities to non-i.i.d.\ training distributions. We will show that we can mitigate forgetting via localization because only fetched value codes are being updated.
Theoretically (see \Cref{sec:analysis}), this is enabled via the following two mechanisms: first, having these intermediate discrete codes minimizes the nonstationarity of the input distribution shift by keeping the joint features fixed; and second, the discrete bottleneck will reduce the complexity of the hypothesis class through its separate key-value codes.

The fundamental difference to prior work on discrete bottlenecks \cite{van2017neural, razavi2019generating, liu2021discrete} is the introduction of the discrete pairs with two different codes, the first (key codes) being optimized for the encoder and the second  (value codes) being optimized for the decoder. This means that the gradients from a particular downstream task will not directly affect the key codes, which induces a rather different learning behaviour compared to standard neural networks. 

\textbf{Contributions:} We introduce a model architecture designed for learning under input distribution changes during training. We corroborate this theoretically by proving that under input distribution shifts, the proposed model achieves an architecture-dependent generalization bound that is better than that of models without the discrete bottleneck. We empirically verified the method under a challenging class-incremental learning scenario on real-world data and show that the proposed model --- without any task boundaries --- reduces the common vulnerability to catastrophic forgetting across a wide variety of publicly available pre-trained models and outperforms relevant baselines on this task. 

\section{Encode, Processing via Discrete Key-Value Bottleneck, Decode}
\label{sec:method}
This section introduces the proposed model architecture. An overview of the workings of the model is given in \Cref{fig:fig1}. Our goal is to learn a model $f_\theta: \mathcal{X} \xrightarrow{} \mathcal{Y}$ from training data $S=((x_i, y_i))_{i=1}^n$ that is robust to strong input distribution changes.
We consider a model that can be decomposed as
\begin{align}
    f_\theta(x) = (d_\delta \circ v_\gamma \circ k_\beta \circ z_\alpha)(x)
\end{align}
with $\theta = (\alpha, \beta, \gamma, \delta)$. In the first step, the encoder $z_\alpha: \mathcal{X}\xrightarrow{}\mathcal{Z} \in \mathbb{R}^{m_z}$ extracts a representation from the high-dimensional observation $x$. We further project this representation into $C$ lower-dimensional feature heads, each of them being passed as the input into a separate head-specific learnable key-value codebook. As the name suggests, a key-value codebook is a bijection that maps each key vector to its value (where the latter is learned). %As we discuss later in \Cref{sec:discussion}, such a decomposition allows us to increase the model's capacity and representative power. 
Within each codebook, a quantization process $k_\beta$ selects the closest key to its input head (think of it as a “feature anchor”) from a head-specific key-value codebook. Next, a corresponding value vector is being fetched and returned (one per codebook), which we denote via $v_\gamma$. The returned values across all codebooks finally serve as input to a decoder $d_\delta$ for prediction. In the following, each component is motivated in detail.\looseness=-1

\textbf{Step 1: Encoding.} 
The encoder $z_\alpha$ projects the input $x$ into a lower-dimensional vector $z \in \mathbb{R}^{m_z}$, further down-projected into $C$ separate heads of dimension $d_{key}$ using $C$ fixed Gaussian random projection matrices. For tasks where we judge $x$ to be sufficiently low-dimensional, we skip encoding and partition $x$ directly. For the purpose of this work we take a broad spectrum of differently pre-trained encoders \cite{caron2021emerging, caron2020unsupervised, he2016deep, trockman2022patches, radford2021learning, dosovitskiy2020image}, and show that they are all capable of embedding features that are relevant for the studied task.

\textbf{Step 2: Processing via Discrete Key-Value Bottleneck.} 
\begin{enumerate}[topsep=0pt,leftmargin=5pt]
    \item[] \textbf{Step 2.1: Quantization.} A quantization process snaps each head to the closest key from the corresponding head's codebook, based on the head's content. Closeness is defined by the L2 distance between the head and key vectors.
    \item[] \textbf{Step 2.2: Retrieval.} For each head, a simple lookup in the head-specific codebook is performed to retrieve the value corresponding to the key selected in the previous step.
\end{enumerate}
Overall, the bottleneck will retrieve a set of $C$ value codes. \looseness=-1

\textbf{Step 3: Decoding.} Finally, we predict the target variable from this set of fetched values using any suitable decoder function $d_\delta$. For the purpose of our experiments on classification, we can apply a simple non-parametric decoder function which uses average pooling to calculate the element-wise average of all the fetched value codes and then applies a softmax function to the output.\looseness=-1

\textbf{Initializing the Model.} During training, we first initialize the key codes. The non-parametric 1-to-1 mapping between key and value codes means no gradient back-propagates from the values to the keys. Instead, they are solely determined from the encoding of the input observation, implying that no supervised data is required to learn them. A simple and effective way to initialize the keys is via exponential moving average (EMA) updates as proposed for VQ-VAE in \cite{van2017neural} and used for VQ-VAE2 in \cite{razavi2019generating}. This procedure allows us to obtain keys that are broadly distributed in the feature space of the encoder, without introducing additional parameters. We refer to \Cref{sec:implementation_details,app:key-analysis} and \Cref{subsec:ablation} for more details on the implementation and robustness of this procedure. 
Once initialized, the keys are frozen and thus not influenced by the task on the target domain, implying that any given input $x$ is mapped to keys from a fixed set.  

\textbf{Training under Dataset Shifts.} 
Assume we are now given training data $S=((x_i, y_i))_{i=1}^n$, but the data generation process might undergo various covariate shifts of $P(X)$ without knowing when they occur. 
A particularly challenging example of that would be the task of continual learning class-incrementally without having task boundaries, where the support of $P(X)$ changes dramatically.
Having the encoder and decoder solely connected through the key-value codes, we can now train the model on this data stream by only updating the value codes under such an input distribution shift.
Since we only locally update the actual retrieved values under the input distribution changes, the values from other data domains remain unchanged, thereby enabling \textit{localized, context-dependent model updates}.
In this way, we can integrate new data and thus gradually improve the models' performance. Moreover, as we will show theoretically, the proposed architecture benefits from the fact that the decoder works with a discrete set of value codes, as opposed to directly predicting from the encoder representation.\looseness=-1

\section{Theoretical Analysis}
\label{sec:analysis}
In this section, we theoretically investigate  the behavior of the proposed method. In contrast to standard models,  the proposed model is shown to have the ability to reduce the generalization error 
 of transfer learning under input distribution shifts via two mechanisms: (1) the discretization with key-value pairs, mitigating the effect of the input distribution shift, and (2)\
the discrete bottleneck with key-value pairs, reducing the complexity of the hypothesis class.\looseness=-1

To formally state this, we first introduce the notation. We consider the distribution shift of $x$ via an arbitrary function $(x, \epsilon) \mapsto g_{\epsilon}(x)$ with random vectors $\epsilon$. In the target domain, we are given a training dataset $S=((x_i,y_i))_{i=1}^n$ of  $n$ samples where $x_i \in \Xcal \subseteq  \RR^{m_x}$ and $y_i \in\Ycal\subseteq  \RR^{m_y}$ are the $i$-th input and the $i$-th target respectively. The training  loss per sample is defined by 
$ \ell(f_{\theta}(x_{i}),y_{i}), 
$
where $\ell: \RR^{m_{y}} \times\Ycal \rightarrow [0,M]$ is a bounded loss function. 
We compare the proposed model against an arbitrary model  $\hf: \Xcal \rightarrow \RR^{m_y}$ without the key-value discretization with a hypothesis class $\hFcal \ni \hf$. 
Let $f_{\theta}^S$ (for the proposed model) and $\hf^S$ (for an arbitrary model) be the corresponding hypotheses learned through the training set $S$. 
Let $\Kcal=\{(k_\beta \circ z_\alpha)(x) : x \in \Xcal \}$ be the set of keys. By ordering elements of $\Kcal$, we use $\Kcal_i$ to denote $i$-th element of $\Kcal$. Let $\hat d$ be a distance function. We define $[n]=\{1,\dots,n\}$,  
$
\Ccal_k=\{x\in \Xcal :k= \argmin_{i \in [|\Kcal|]}  \hat d((k_\beta \circ z_\alpha)(x),\Kcal_{i}) \}
$ (the input region corresponding to the $k$-th key), and
$\Ical_{k}^{y}=\{i\in[n]: x_i \in \Ccal_{k}, y_i=y\}$ (the data indices corresponding to the $k$-th key and the label $y$).
We define $\one\{a=b\}=1$ if $a=b$ and $\one\{a=b\}=0$ if $a\neq b$. We denote by $\Dcal_{x|y}$ the conditional distribution of the input $x$ given a label $y$, and by $\bar S=((\bx_i,\by_i))_{i=1}^n$ the random variable having the same distribution as $S$.

The following theorem (proof in \Cref{sec:proof}) shows that under input distribution shifts, a model with the proposed architecture has the ability to reduce the generalization error of the base model:\\
\begin{theorem} \label{thm:1}
There exists a constant $c$ (independent of $n,f,\hFcal,\delta, \epsilon$ and $S$) such that for any $\delta>0$,  with  probability at least $1-\delta$, the following holds for any $f_{}\in \{f_{\theta}^{S}, \hf^S\}$: 
\begin{align*}
%\small
\EE_{(x,y),\epsilon}[\ell(f(g_\epsilon(x)),y_{})]  \le \frac{1}{n} \sum_{i=1}^n \ell(f(x_{i}),y_{i}) \\
 +  c  \sqrt{\frac{2\ln(2e/\delta)}{n}} +\Bcal_{g_{\epsilon}}(f)+ \one\{f=\hf^S\}\Gcal_{\hf^S},
\end{align*} 
where 
\begin{equation*}
\begin{split}
\Bcal_{g_{\epsilon}}(f)=\frac{1}{n}\sum_{y,k}|\Ical_{k}^{y}| \cdot \EE_{x\sim \Dcal_{x|y},\epsilon}\left[\ell(f(g_\epsilon(x)),y_{}) \right.\\ 
- \left. \ell(f(x),y_{})|x\in  \Ccal_{k} ^{}\right],
\end{split}
\end{equation*}
and $\Gcal_{\hf^S}= \sum_{y,k}\frac{2 \Rcal_{y,k}(\ell \circ \hFcal)}{n}+c  \sqrt{\frac{\ln(2e/\delta)}{2n}}$. Here, with independent uniform random variables $\xi_{1},\dots,\xi_{n}$ taking values in $\{-1,1\}$, $\Rcal_{y,k}(\ell \circ \hFcal)=\EE_{\bar S,\xi} [\sup_{\hf \in\hFcal} \sum_{i=1}^{|\Ical_{k}^{y}|}  \xi_{i}\ell(\hat f(\bx_{i}),\by_{i}) |\bx_{i}\in  \Ccal_{k},\by_{i}=y]$. Moreover,  if  $x\in \Ccal_k \Rightarrow g_\epsilon(x) \in \Ccal_k$ with probability one, then 
$
\Bcal_{g_{\epsilon}}(f_{\theta}^{S})=0.
$
\end{theorem}

\begin{figure*}[ht]
    \centering
    \includegraphics[width=1.0\textwidth]{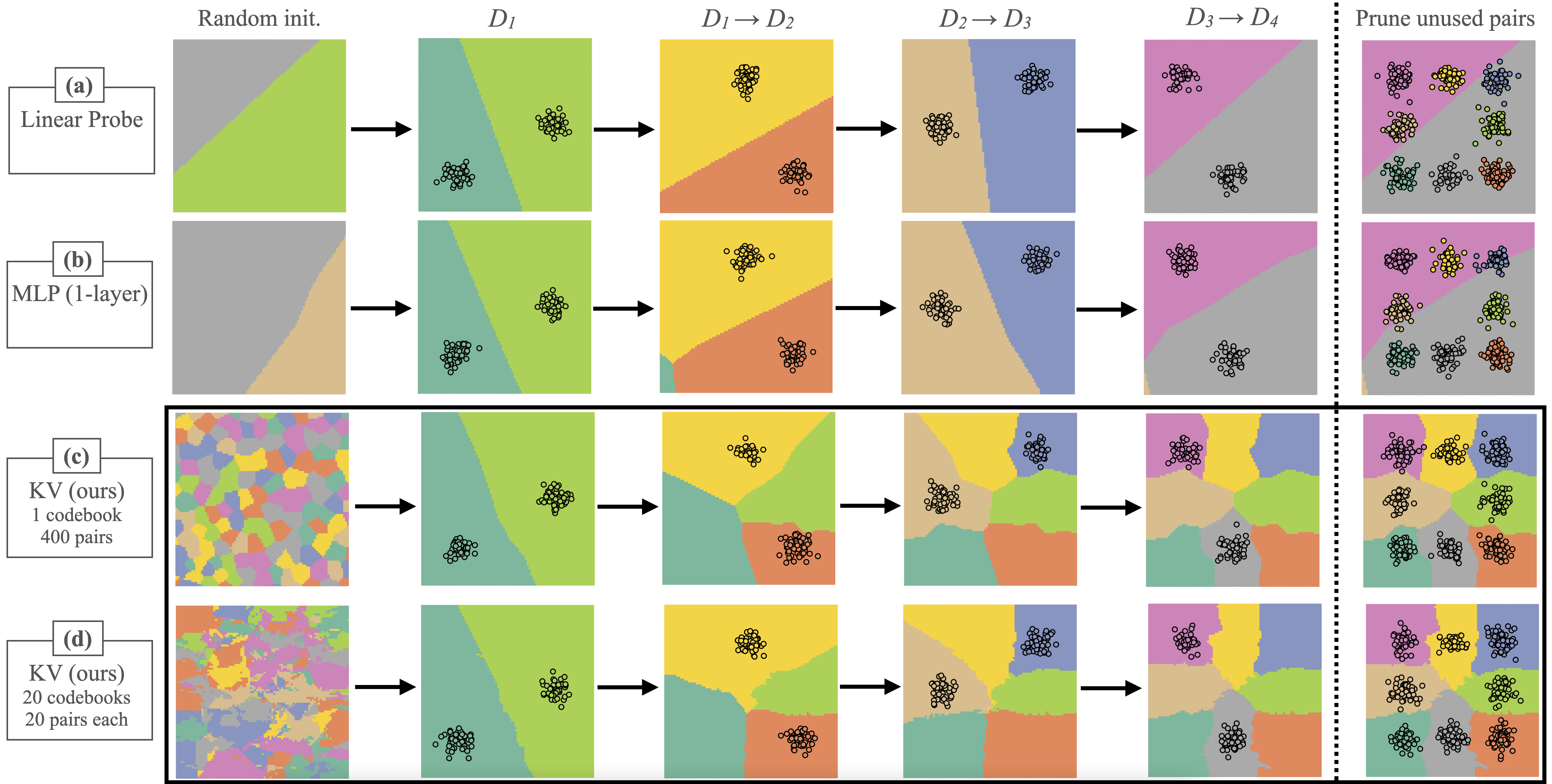}
    \caption{
    Incremental training on non-overlapping datasets $\mathcal{D}_1 \xrightarrow{} \mathcal{D}_2 \xrightarrow{} \mathcal{D}_3 \xrightarrow{} \mathcal{D}_4$ (dots) using (a) Linear Probe (LP), (b) 1-layer MLP (MLP), (c) key-value bottleneck with C = 1 and ground truth feature heads, (d) key-value bottleneck with C = 20 and randomly projected feature heads. Using a discrete bottleneck of 400 key-value pairs with fixed key codes, and locally updated value codes, alleviates catastrophic forgetting. Color domains indicate one out of the 8 possible classes predicted by the model.\looseness=-1
    }
    \label{fig:fig2}
\end{figure*}

\textbf{Implications for Empirical Results:} 
The first benefit of the use of key-value pairs appears in the training loss $\frac{1}{n} \sum_{i=1}^n \ell(f(x_{i}),y_{i})$ in the right-hand side (RHS) of the generalization error bound in \Cref{thm:1}. Since the  bound is independent of the complexity of $v_\gamma$, we can set the hypothesis class of the local values $v_\gamma$ to be highly expressive to  reduce the training loss for each local region of the input space (in order to minimize the RHS of the  bound in Theorem \ref{thm:1}). Indeed, \Cref{fig:fig2} demonstrates this (although the main purpose is to show another benefit in terms of localization and forgetting). 
Moreover, the benefit of the proposed architecture is captured by two further mechanisms:
The first mechanism, the term $\Bcal_{g_{\epsilon}}(f)$ in Theorem \ref{thm:1} measures the effect of the input distribution shift $g_{\epsilon}$. Theorem \ref{thm:1} shows that if the key of $x$ is still the key of $g_{\epsilon}(x)$ (i.e., $x\in \Ccal_k \Rightarrow g_\epsilon(x) \in \Ccal_k$) with probability one, then we can further reduce the error as $\Bcal_{g_{\epsilon}}(f_{\theta}^{S})=0$ while $\Bcal_{g_{\epsilon}}(\hf^{S})>0$ in general. This captures the mechanism, which is the minimization of the effect of the input distribution shifts via the discretization with key-value pairs.
The second mechanism is shown by the fact that a model with the proposed architecture reduces the bound by making the last term $\one\{f=\hf^S\}\Gcal_{\hf^S}$ vanish, because $\one\{f=\hf^S\}=0$ for $f=f^S_\theta$. This removed term consists of the complexity of  its hypothesis class $ \Rcal_{y,k}(\ell \circ \hFcal)$. Thus, the proposed model avoids the complexity cost as compared to other architectures. This is because the \textit{bottleneck} with key-value pairs prevents over-fitting. 
Finally, the benefit from random projections in terms of \Cref{thm:1} is that this projection can reduce the term of $\Bcal_{g_{\epsilon}}(f)$ in addition to the flexibility in reducing the training loss term when having multiple codebooks as further discussed in \Cref{app:random-matrix}.\looseness=-1

\section{Related Work}
The hallmark of machine learning is to be able to develop models that can quickly adapt to new tasks once trained on sufficiently diverse tasks \cite{baxter2000model, thrun2012learning}. Multiple ways to transfer information from one task to another exist: (1) transfer information via the transfer of the neural network weights (when trained on source tasks); (2) reuse raw data as in retrieval-based methods \cite{borgeaud2021improving, nakano2021webgpt,lee2019latent, lewis2020retrieval, guu2020realm, sun2021ernie, goyal2022retrieval}; or (3), via knowledge distillation \cite{hinton2015distilling}. Each approach implies inevitable trade-offs: When directly transferring neural network weights, previous information about the data may be lost in the fine-tuning process, while transfer via raw data may be prohibitively expensive as there can be hundreds of thousands of past experiences.
At the same time, we want models that can continually accumulate information without training from scratch and forgetting previous information  \cite{chen2018lifelong, thrun1995learning, davidson2020sequential}. To prevent models from forgetting previous information, continual learning approaches make use of replay-buffers to replay old information \cite{rebuffi2017icarl, li2017learning, shin2017continual, tang2022learning}, regularization during optimization \cite{kirkpatrick2017overcoming, zenke2017continual}, incrementally adding task-specific model components to increase capacity \cite{xiao2014error}, using parameter isolation-based approaches \citep{verma2021efficient, gao2022efficient, serra2018overcoming, ke2021achieving}, or meta-learning fast-adapting models to promote quick adaptation \cite{harrison2020continuous, he2019task, finn2019online}.  
The proposed method can be mostly linked with parameter isolation-based approaches by isolating the key-value pair for different tasks and classes.
This bears some similarity with the theory of Sparse Distributed Memory (SDM), which is a model of associative memory inspired by human long-term memory and connections to neural circuits found in various organisms \citep{kanerva1988sparse, kanerva1992sparse}. 
The key-value bottleneck method has some similarities with the concurrent work by \citet{sdm_paper} which builds upon SDM ideas, but there are several crucial differences.
While this model also writes and reads from a fixed set of "neurons",
the proposed method does not require the implementation and careful tuning of additional neurobiology-inspired model components, such as "GABA switch implementations" to avoid "dead-neuron problems". 
Furthermore, key-value bottlenecks benefit from handling high-dimensional embedding spaces through random down-projection matrices and multiple distributed codebooks, while maintaining theoretical guarantees on catastrophic forgetting, thereby performing better on the presented task.
The benefit of random down-projections is further supported by recent work exploring the effect of orthogonal projection-based methods as weight regularization to avoid forgetting \citep{zeng2019continual, saha2021gradient}. 
Other models such as \citet{shen2021algorithmic} and \citet{pourcel2022online} focus on the online-learning setting without being exposed to the same classes for thousands of epochs.
Most approaches for continual learning require the provision of task boundaries, but the proposed model architecture using discrete key-value codes avoids this constraint. The information is distilled into discrete key-value codes and a sparse set of value codes are locally adapted when faced with new tasks, which integrates information organically as discussed in \Cref{sec:analysis} and has some parallels with local learning ideas \citep{bottou1992local}. A central part of the proposed method is the use of pre-trained models, a prerequisite that is also harnessed for continual learning tasks recently \cite{banayeeanzade2021generative, ostapenko2022foundational}. 
The proposed method builds upon discrete representations, most notably the VQ-VAE \cite{van2017neural, liu2021discrete} and various successor methods that have been shown to improve robustness under data corruptions \cite{razavi2019generating, liu2021discrete, zeghidour2021soundstream, yu2021vector, shin2021translation, mama2021nwt}. While these works can similarly reduce the representation space and enhance robustness, such bottlenecks have a single set of codes, which does not allow tuning them while keeping the feature-extracting backbone frozen.
Finally, the key-value pairs can be interpreted as a memory bottleneck from which the model can selectively retrieve stored information. This interpretation bears similarity to memory networks, such as those proposed by \cite{sukhbaatar2015end, chandar2016hierarchical, webb2020emergent, lample2019large, panigrahy2021sketch}, which also write and read information from a set of symbol-like memory cells. However, key-value codes act as an information bottleneck, whereas memory networks condition predictions on retrieved memory. \looseness=-1

\section{Experiments}
\label{sec:experiments}

The insight we aim to convey with the presented experiments is to validate that having a key-value bottleneck allows for \textit{localized, context-dependent model updates} that are robust to training under input distribution changes. We will show that this allows us to \textit{reduce catastrophic forgetting}; \textit{integrate new information}; and thereby \textit{mitigate vulnerabilities} in challenging class-incremental learning settings. We first showcase a simplified yet challenging class-incremental learning setting to give readers an understanding of why standard approaches fail and how the proposed method succeeds. We then explore the method's benefits under a challenging class-incremental learning scenario on CIFAR10 and analyze the role of different components and design choices of a key-value bottleneck. The code for reproducing all experiments can be found in the supplementary material.

\subsection{A Simple Learning Setting Motivating the Method}
We consider a 2D input feature classification problem for 8 classes, where the training data is not i.i.d.\ but changes over four stages (see \Cref{fig:fig2}). In each stage, we sample 100 examples of two classes for 1000 training steps, using gradient descent to update the weights, then move on to two new classes for the next 1000 steps. The input features of each class follow spatially separated normal distributions.
We first discuss the behaviour on this task using a vanilla linear probe or 1-layer MLP with 32 hidden dimensions trained by minimizing the cross-entropy along the training data stream. As can be seen from the decision boundaries in \Cref{fig:fig2} (a, b), these naive approaches --- unsurprisingly --- over-fit on the most recent training data. This behaviour is because the MLP is optimizing \textit{all} weights on the most recent training data, thereby overwriting prior predictions. This problem can be easily alleviated by \textit{sparse, local} updates using a discrete key-value bottleneck. Here we present two different configurations. In the first configuration, we use 400 pairs of 2-dim keys and 8-dim values without random projection and we initialize these keys from random points in the domain $x \in [0, 1]^2$ (not the target dataset). 
Input $x_i$ will now snap to the closest key and fetch its corresponding value. After keys are initialized, they are frozen, and we train the model on the non-stationary data stream using the same loss and optimizer as before but optimizing the values only. As can be seen from \Cref{fig:fig2} (c), the proposed bottleneck mechanism will now solely update the input-dependent value codes. This enables the model to incrementally update its predictions on unseen input domains and minimize interference. Finally, we can prune all key-value pairs that were never selected in order to remove any dormant key-value codes (rightmost column). In the above example, it is essential to have two separate sets of codes compared to a single set of discrete codes. Finally, we also study a configuration with 20 codebooks each with 20 key-value pairs and use randomly initialized projection matrices. 
This configuration, which uses multiple codebooks but the same total number of key-value pairs, works just as well as seen in \Cref{fig:fig2} (d).\looseness=-1

\begin{figure*}[ht]
    \centering
    \includegraphics[width=1.0\textwidth]{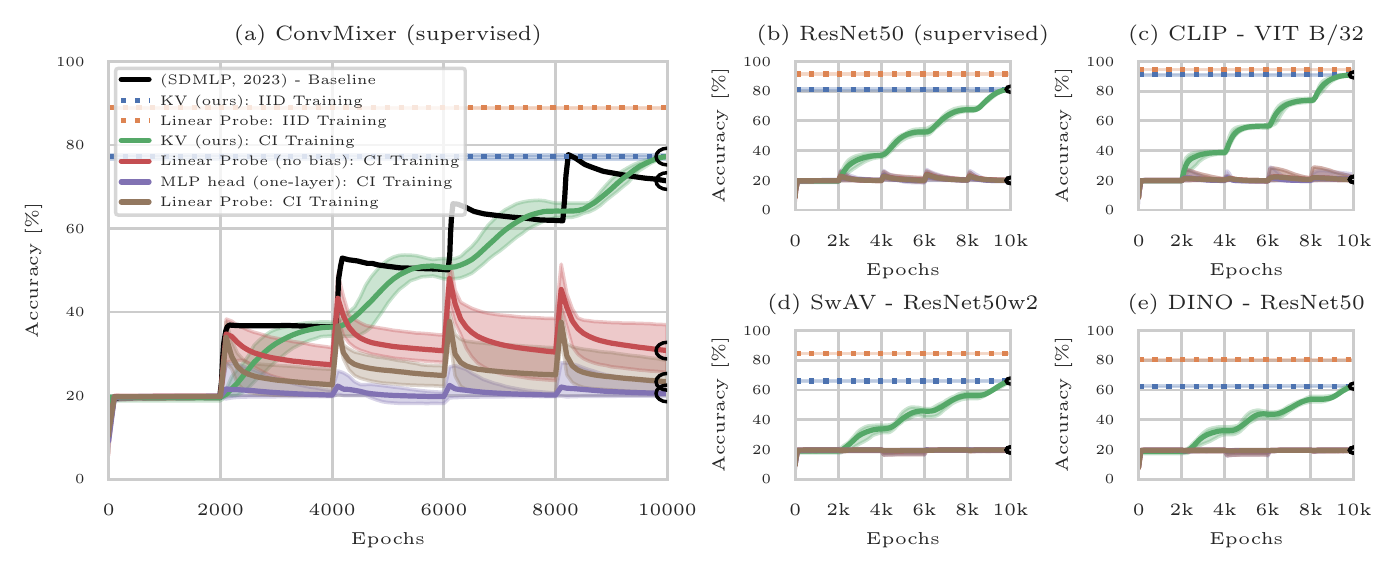}
    \caption{
    The key-value bottleneck architecture successfully learns and integrates new information in a challenging 5-split CIFAR10 class-incremental (CI) learning setting. Our method (green line) reduces catastrophic forgetting and yields strong final performance across various pre-trained frozen backbone architectures including (a) ConvMixer; (b) ResNet50 ; (c) ViT B/32 with CLIP, (d) ResNet50w2 with SwAV; (e) ResNet50 with DINO. At the end of the training, the proposed method outperforms the reported performance of the SDM method using the same pre-trained ConvMixer. Standard probe tuning exhibits major forgetting in this scenario (red, brown, purple curve).\looseness=-1
    \looseness=-1}
    \label{fig:fig3}
\end{figure*}

\subsection{Continual Learning on Real-World Data}
Having established some first intuition on the localized model updates, we now attempt to present the strong learning capabilities of the introduced bottleneck on a challenging and realistic continual class-incremental learning task. 

\textbf{Experimental Setup.}
We use a class-incremental CIFAR10 task with pre-training on another dataset without any memory replay or provision of task boundaries, which is a very challenging task setting in continual learning. Here, five disjoint sets with two classes each are incrementally presented for many epochs each and the goal is to learn new classes and not forget previous classes at the very end of the training. We present each set for 2000 epochs, which in an ordinary architecture would cause catastrophic forgetting of the previous sets. We perform five replications of each model with different random seeds, with the seed also re-sampling the class splits to avoid selecting any favourable class split. Importantly, we study one of the hardest yet most realistic settings in continual learning by \emph{not allowing any memory replay or provision of task boundaries} which is required by the vast majority of existing continual learning methods. To the best of our knowledge, the only method that can deal with this learning scenario is \citep{sdm_paper}.
We report experiments on five publicly available backbones including:\looseness=-1
\begin{enumerate}
\item ResNet50 pre-trained on ImageNet
\item ViT-B/32 pre-trained with CLIP
\item ResNet50w2 pre-trained with SwAV as SSL method 
\item ResNet50 pre-trained with DINO as SSL method 
\item ConvMixer pre-trained on 32x32 downsampled Imagenet. We select this backbone to directly compare against results reported in \citep{sdm_paper}. 
\end{enumerate}
We initialize keys on the unlabelled non-overlapping CIFAR100 dataset except for the ConvMixer where we used the embeddings from the downsampled Imagenet dataset for reasons of comparison. 
We test the wide applicability of key-value bottlenecks across these different architectures and pre-training schemes, using 256 codebooks with 4096 key-value pairs each, 14-dimensional keys and 10-dimensional values. We additionally report results for the proposed bottleneck and a Linear Probe under an IID training scheme (presenting all classes simultaneously), with the latter representing a fair oracle upper bound. 
When learning class-incrementally, we compare against the reported SDMLP performance on the ConvMixer backbone, as well as a 1-layer MLP with 128 hidden dimensions and a linear probe with and without bias term for all backbones. We refer to \Cref{app:additional-experiments} for further experiments.

\textbf{Results.}
We summarize our results in \Cref{fig:fig3}. First, we observe that the key-value bottleneck can indeed successfully learn and integrate new information under such a difficult class-incremental distribution shift consistently (green curves). At the same time, the key-value bottleneck substantially reduces the severe catastrophic forgetting over the standard adaptation approaches (red, brown and purple curves). In line with the theory, we observe \textit{across all encoder backbones} the appealing learning property that the bottleneck yields the same performance at the end of training compared to the benign i.i.d.\ setting (blue lines). Within the class incremental setting only, we observe that the \textit{final performance} is almost identical, irrespective of the sampled class split. This result suggests that the bottleneck can successfully mitigate the typically very destructive effect of learning under non-stationary training data and might be particularly robust against training curricula that are otherwise adversarial to stochastic gradient-based training. 
We can partially explain this behaviour by the fact that we freeze the keys during training and use a simple momentum-free SGD optimizer. In this case, any value code will approximatively receive the same gradients over the entire course of training.  

\begin{figure}
\centering
    \includegraphics[width=\columnwidth]{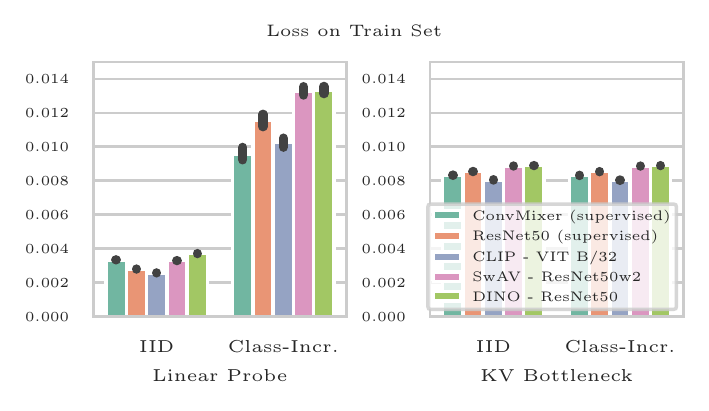}
    \vspace{-0.5cm}
    \caption{We demonstrate that the proposed key-value bottleneck method maintains a constant loss measure, directly translating to improved generalization when training class incrementally, as opposed to standard approaches which can over-fit.
    \looseness=-1}
    \label{fig:fig3b}
\end{figure}

We now take a look at the results of the ConvMixer backbone in \Cref{fig:fig3} (a). This reflects the same setup as in the SDMLP method, the only applicable method we are aware of in this challenging task setting without memory replay or provision of task boundaries. The key-value bottleneck method achieves a final accuracy of 77.3\% compared to 71\% for the best final accuracy with SDMLP. 
Additionally, the learning curves with the bottleneck show smooth improvements in performance, whereas the baseline spikes at the introduction of new tasks and exhibits pronounced catastrophic forgetting.
Furthermore, we observe consistent results across all other four backbones. The best results are achieved with a CLIP-encoder with 91.3\% and the supervised pre-trained ResNet50 backbone with 81.5\% final performance (\Cref{fig:fig3} b \& c). We are similarly getting strong class-incremental learning abilities for both SSL backbones, which were pre-trained without any labels, achieving up to 66.3\% in the case of SwAV and 62.5\% in the case of DINO (\Cref{fig:fig3} d \& e).
Finally, we believe these results are further encouraging as they demonstrate that the key-value bottleneck is able to achieve strong performance across a wide variety of pre-trained backbones and learned feature spaces, suggesting that it has the potential to scale to even more performant models in the future.
Finally, we show in \Cref{fig:fig3b} that a key-value bottleneck does not over-fit like other standard approaches. A linear probe may perform better when trained on all classes at once in an i.i.d.\ setting; but when trained incrementally, it over-fits. However, the key-value bottleneck maintains a constant loss measure over both, i.i.d.\ and class-incremental settings, which leads to better overall performance when training on new classes. 
This aligns with our analysis of the first term in \Cref{thm:1}.

\begin{figure*}[t]
    \centering
    \includegraphics[width=\textwidth]{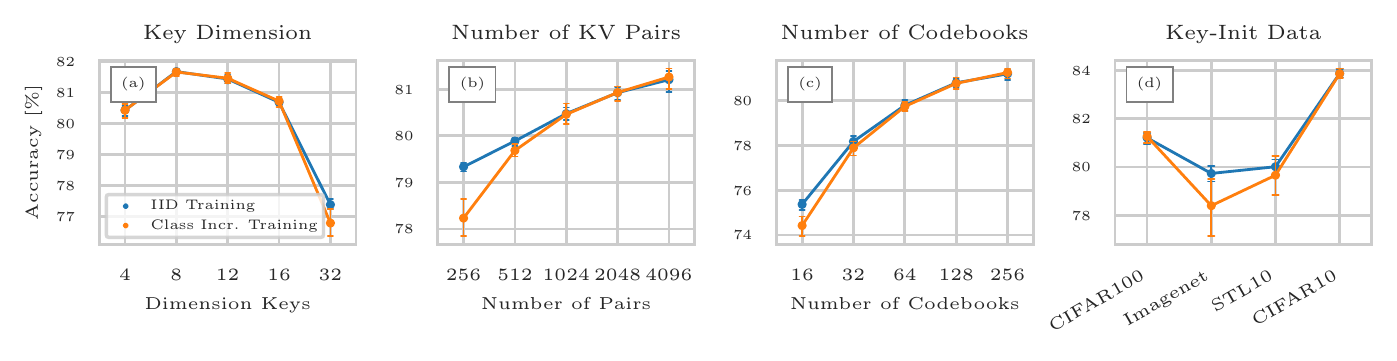}
    \caption{Assessing the role of individual model components starting from the base model architecture in \Cref{fig:fig3} b. Reducing the dimensionality of keys has a positive effect (a); increasing the number of pairs and codebooks enhances generalization performance (b + c); 
    finally, key-value bottlenecks keys can be initialized from a different dataset without major drop in overall performance (d).\looseness=-1}
    \label{fig:fig4}
\end{figure*}

\subsection{Ablation and Sensitivity Analysis}
\label{subsec:ablation}
We now analyze the role of various relevant bottleneck components to better understand its learning abilities, as well as further opportunities and limitations. We point the reader to \Cref{app:key-analysis} for an in-depth analysis of the (key,value) codes learned by the models. In the following paragraphs, we focus on the ResNet50 backbone and analyze the effect of changing one particular component while keeping everything else fixed. We present the results in \Cref{fig:fig4}.

\textbf{Dimension of Key Codes.} 
First, we found that there is an optimal number of dimensions for the keys, around 8 to 12, which can be explained as follows. Having keys with high dimensions requires more data to cover the backbone manifold sufficiently. On the other hand, having keys with too few dimensions means that the keys are scattered in a much smaller space which increases the chances of unintended key sharing among different classes.

\textbf{Number of Key-Value Pairs.} The number of keys used in the bottleneck can greatly affect the performance of the model. We found that increasing the number of key-value pairs leads to better performance, with only a small decrease in performance when using four times fewer pairs. Our analysis also suggests that the model performance will continue to improve asymptotically, though we have not yet reached that point. When using 256 or fewer key-value pairs, we noticed a separation between the performance of the model in i.i.d.\ and class-incremental settings. This is likely because when there are fewer pairs, more of them are shared among multiple classes, and this allows for a more balanced update of the gradients when training on mini-batches.

\textbf{Number of Codebooks.} The number of codebooks is another important factor that controls the capacity of the model, similar to the number of key-value pairs. We observed that using more codebooks leads to better performance, with a slightly more pronounced effect compared to the number of pairs. Furthermore, increasing the number of codebooks also decreases the training loss term. However, it is important to note that increasing the number of codebooks too much can also increase the value of $c$ in \Cref{thm:1}, although we have not yet observed that in our experiments.

\textbf{Key Initialization Dataset.} Finally, we tested the robustness of the proposed method by initializing the keys on different datasets, such as STL10 and Imagenet, in addition to CIFAR100 used in previous experiments \citep{coates2011analysis, krizhevsky2009learning}. We also calculated the performance of an "oracle" key initialization by using the unlabelled CIFAR10 dataset. While there is a slight decrease in performance compared to the oracle, the architecture is robust to different datasets, even if
those datasets differ in resolution and scenes.\looseness=-1

\section{Limitations and Extensions}
\label{sec:discussion}

\textbf{Pre-trained Encoders.} The key-value bottleneck method relies on pre-trained encoders that can extract meaningful features shared between the pre-training and actual training data. This is an important requirement that has been shown by prior works in transfer learning \cite{yosinski2014transferable, chen2020big, azizi2021big}. However, as self-supervised pre-trained models from massive datasets become more prevalent, a key-value bottleneck can be used to connect these encoders to downstream tasks and minimize forgetting under distribution changes. Additionally, the method relies on encoders that are able to separate distinct features locally, as supported by recent research \cite{huang2019unsupervised, caron2021emerging, zhou2022understanding, tian2022deeper}. Fine-tuning the encoder directly can distort these features and, therefore, negatively affect out-of-distribution generalization \cite{kumar2022fine}. The key-value bottleneck reduces this issue as encoder features are not influenced by fine-tuning.\looseness=-1

\textbf{Selection and Initialization of Key Codes.} 
The selection and initialization of key codes is an important aspect of the proposed method. The approach of using EMA works well for a range of datasets, but it may have limitations in extreme situations where the dataset is unknown or changes completely, such as in RL. We did not explore these scenarios in this work, but we believe that discrete key-value bottlenecks have the potential to be extended to handle them. For example, the encoder, projection, and keys can be fine-tuned asynchronously with auxiliary unsupervised tasks to handle distribution shifts. Additionally, information about the distance between the codebook input heads and their closest keys can be used to estimate the level of distribution shift and adapt, add, or reinitialize key-value pairs as needed.\looseness=-1

\textbf{Decoder Architecture} 
When classifying, we found it effective to use a non-parametric decoder without learnable weights. Nevertheless, we can also train a discrete key-value bottleneck with more complex, parametric decoders. When training the model under i.i.d.\ conditions, we can adjust the decoder weights to improve performance. However, when the training conditions change, e.g.\ in incremental learning, we need to be more careful. One could make the decoder weights adjustable and augment with various regularization methods or keep them fixed if necessary. Combining key-value bottlenecks with advanced, pre-trained decoders such as generative models is a promising area for future research.

\textbf{Trade-offs using a Bottleneck.} Using an information bottleneck involves a trade-off between accuracy and information compression.
The proposed bottleneck applies strong compression by propagating only a discrete set of C key-value pairs, preventing changes to the encoder.
This allows for training under strong distribution shifts without forgetting, but with slightly lower performance.
If the target data distribution is available before (no labels needed), we can improve performance by initializing the keys on this distribution, but there are also other ways to potentially improve further such as parametric decoders or allowing back-propagation into the projection weights and keys through techniques such as sampling or straight-through estimators. 

\section{Conclusion}
We proposed a discrete key-value bottleneck architecture that can adapt to input distribution shifts and performs well in class-incremental learning settings. We supported the model's effectiveness theoretically and through empirical validation. Specifically, we showed that the use of key-value pairs in the architecture can reduce the generalization error and prevent over-fitting. The model is able to integrate new information locally while avoiding catastrophic forgetting. We believe the key-value bottleneck could be a promising solution for tackling challenging training scenarios.\looseness=-1

\section*{Acknowledgements}
We would like to thank Jonas Wildberger, Andrea Banino, Andrea Dittadi, Diego Agudelo España, Chiyuan Zhang, Arthur Szlam, Olivier Tieleman, Wisdom D'Almeida, Felix Leeb, Alex Lamb, Xu Ji, Dianbo Liu, Aniket Didolar, Nan Rosemary Ke and Moksh Jain for valuable feedback. The authors thank the International Max Planck Research School for Intelligent Systems (IMPRS-IS) for supporting FT.

\bibliography{icml2023}
\bibliographystyle{icml2023}

\clearpage

\onecolumn

\appendix
\onecolumn

\section{Proof of Theorem \ref{thm:1}}
\label{sec:proof}
\begin{proof}
Let  $f_{}\in \{f_{\theta}^{S}, \hf^S\}$. We define 
$$
\Ccal_{k}^{y}=\left\{(\bx ,\by)\in \Xcal\times \Ycal  : y=\by, k= \argmin_{i \in [|\Kcal|]}  d((k_\beta \circ z_\alpha)(\bx),\Kcal_{i}) \right \}.
$$ 
We use the notation of $z = (x, y)$ and $\tx^\epsilon=g_\epsilon(x)$. We first write the expected error as the sum of the conditional expected error:
\begin{align*}
\EE_{z}[\ell(f(\tx^\epsilon),y_{})] &=\sum_{k,y} \EE_{x}[\ell(f(\tx^\epsilon),y_{})|z\in  \Ccal_{k}^{y}]\Pr(z_{}\in  \Ccal_{k}^{y}) =\sum_{k,y} \EE_{x_{k,y}}[\ell(f(\tx^\epsilon_{k,y}),y_{})]\Pr(z\in  \Ccal_{k}^y),
\end{align*}
where $\tx^\epsilon_{k,y}=g_\epsilon(x_{k,y})$ and $x_{k,y}$ is the random variable $x$ conditioned on $z\in  \Ccal_{k}^y$. 
Using this, we  decompose the generalization error into two terms:
\begin{align} \label{eq:1}
\EE_{z}[\ell(f(\tx^\epsilon),y)]  - \frac{1}{n} \sum_{i=1}^n \ell(f(x_{i}),y_{i})
 & =\sum_{k,y} \EE_{x_{k,y}}[\ell(f(\tx^\epsilon_{k,y}),y)]\left(\Pr(z\in  \Ccal_{k}^{y})- \frac{|\Ical_{k}^{y}|}{n}\right)
\\ \nonumber & \quad +\left(\sum_{k,y}^{} \EE_{x_{k,y}}[\ell(f(\tx^\epsilon_{k,y}),y)]\frac{|\Ical_{k}^{y}|}{n}- \frac{1}{n} \sum_{i=1}^n \ell(f(x_{i}),y_{i})\right). 
\end{align}
Since 
$
\frac{1}{n} \sum_{i=1}^n \ell(h, r_i)=\frac{1}{n}\sum_{y} \sum_{k\in I_{\Kcal}^{y}}  \sum_{i \in \Ical_{k}^y}\ell(f(x_{i}),y_{i}), 
$
the second term in the right-hand side  of \eqref{eq:1} is further simplified 
as 
\begin{align*}
& \sum_{k,y}^{} \EE_{x_{k,y}}[\ell(f(\tx^\epsilon_{k,y}),y)]\frac{|\Ical_{k}^{y}|}{n}- \frac{1}{n} \sum_{i=1}^n \ell(f(x_{i}),y_{i})
\\ & =\frac{1}{n}\sum_{y} \sum_{k\in I_{\Kcal}^{y}}|\Ical_{k}^{y}|\left(\EE_{x_{k,y}}[\ell(f(\tx^\epsilon_{k,y}),y)]-\frac{1}{|\Ical_{k}^{y}|}\sum_{i \in \Ical_{k}^y}\ell(f(x_{i}),y_{i}) \right).
\
\end{align*}
where $I_{\Kcal}^{y}=\{k \in [|\Kcal|]:|\Ical_{k}^{y}| \ge 1 \}$. Substituting these into equation \eqref{eq:1} yields
\begin{align} \label{eq:2} 
\EE_{z}[\ell(f(\tx^\epsilon),y_{})]  - \frac{1}{n} \sum_{i=1}^n \ell(f(x_{i}),y_{i}) 
 & =\sum_{k,y}^{} \EE_{x_{k,y}}[\ell(f(\tx^\epsilon_{k}),y)]\left(\Pr(z\in  \Ccal_{k}^{y})- \frac{|\Ical_{k}^{y}|}{n}\right)
 \\ \nonumber & \quad +\frac{1}{n}\sum_{y} \sum_{k\in I_{\Kcal}^{y}}|\Ical_{k}^{y}|\left(\EE_{x_{k,y}}[\ell(f(\tx^\epsilon_{k,y}),y_{})]-\frac{1}{|\Ical_{k}^{y}|}\sum_{i \in \Ical_{k}^y}\ell(f(x_{i}),y_{i}) \right)
\end{align}
By using Lemma 1 of \cite{kawaguchi2022robustness}, we have that for any $\delta>0$, with probability at least $1-\delta$,
\begin{align} \label{eq:3} 
&\sum_{k,y} \EE_{x_{k,y}}[\ell(f(\tx^\epsilon_{k}),y)]\left(\Pr(z\in  \Ccal_{k}^{y})- \frac{|\Ical_{k}^{y}|}{n}\right) 
\\ \nonumber & \le  \left(\sum_{k,y}  \EE_{x_{k,y}}[\ell(f(\tx^\epsilon_{k}),y)] \sqrt{\Pr(z\in  \Ccal_{k}^{y})} \right) \sqrt{\frac{2\ln({|\Ycal||\Kcal|}/\delta)}{n}}
\\ \nonumber & \le M  \left(\sum_{k,y}  \sqrt{\Pr(z\in  \Ccal_{k}^{y})} \right)  \sqrt{\frac{2\ln({|\Ycal||\Kcal|}/\delta)}{n}}.
\end{align}
Here, we have that  $\ln({|\Ycal||\Kcal|}/\delta)=\ln(e^{\ln|\Ycal||\Kcal|}/\delta)\le \ln(e^\zeta/\delta)\le \ln((e/\delta)^{\zeta})=\zeta\ln(e/\delta)$ where $\zeta =\max(1, \ln|\Ycal||\Kcal|)$, since $\delta \in (0,1)$ and $\zeta \ge 1$. Moreover, note that for any $(f,h,M)$ such that $M>0$ and $B\ge0$ for all $X$, we have that 
$
\PP(f(X)\ge M)\ge \PP(f(X)>M) \ge  \PP(Bf(X)+h(X)>BM+h(X)),
$
where the probability is with respect to the randomness of  $X$.
Thus,  by combining \eqref{eq:2} and \eqref{eq:3}, we have that 
for any $\delta>0$, with probability at least $1-\delta$,
 the following holds:
\begin{align} \label{eq:4}
&\EE_{z,\epsilon}[\ell(f(g_\epsilon(x)),y_{})]  - \frac{1}{n} \sum_{i=1}^n \ell(f(x_{i}),y_{i}) 
\\ \nonumber & =\EE_\epsilon\left[\EE_{z}[\ell(f(g_\epsilon(x)),y_{})]  - \frac{1}{n} \sum_{i=1}^n \ell(f(x_{i}),y_{i}) \right] 
\\ \nonumber & \le \frac{1}{n}\sum_{y \in \Ycal} \sum_{k\in I_{\Kcal}^{y}}|\Ical_{k}^{y}|\left(\EE_{z,\epsilon}[\ell(f(g_\epsilon(x)),y_{})|z\in  \Ccal_{k}^{y}]-\frac{1}{|\Ical_{k}^{y}|}\sum_{i \in \Ical_{k}^y}\ell(f(x_{i}),y_{i}) \right)+c  \sqrt{\frac{2\ln(e/\delta)}{n}}
\\ \nonumber & =\frac{1}{n}\sum_{y \in \Ycal}\sum_{k\in I_{\Kcal}^y}|\Ical_{k}^{y}|\left(\EE_{z}[\ell(f(x),y_{})|z\in  \Ccal_{k}^{y}]-\frac{1}{|\Ical_{k}^{y}|}\sum_{i \in \Ical_{k}^y}\ell(f(x_{i}),y_{i}) \right)
\\ \nonumber & \quad +\frac{1}{n}\sum_{y \in \Ycal}\sum_{k\in I_{\Kcal}^y}|\Ical_{k}^{y}|\EE_{z}\left[ \EE_\epsilon[\ell(f(g_\epsilon(x)),y_{})]-\ell(f(x),y_{})|z\in  \Ccal_{k} ^{y}\right] 
 \\ \nonumber & \quad +c  \sqrt{\frac{2\ln(e/\delta)}{n}}
\end{align}
We now bound each term in   the right-hand side of equation \eqref{eq:4} for both cases of $f=\hf^S$ and $f= f^S_\theta$. We first consider the  case of  $f=\hf^S$. For the first term in the right-hand side of equation \eqref{eq:4}, we invoke Lemma 4 of \cite{pham2021combined} to obtain that for any $\delta>0$, with probability at least $1-\delta$,
\begin{align*}
& \frac{1}{n}\sum_{y \in \Ycal}\sum_{k\in I_{\Kcal}^y}|\Ical_{k}^{y}|\left(\EE_{z}[\ell(\hf^S(x),y_{})|z\in  \Ccal_{k}^{y}]-\frac{1}{|\Ical_{k}^{y}|}\sum_{i \in \Ical_{k}^y}\ell(\hf^S(x_{i}),y_{i}) \right) 
\\ & \le\frac{1}{n}\sum_{y \in \Ycal}\sum_{k\in I_{\Kcal}^y}|\Ical_{k}^{y}|\left(2\Rcal_{y,k}(\ell \circ \hFcal)+M \sqrt{\frac{\ln(|\Ycal||\Kcal|/\delta)}{2|\Ical_{k}^{y}|}} \right) 
\\ & =2 \sum_{y \in \Ycal}\sum_{k\in I_{\Kcal}^y}|\Ical_{k}^{y}|\frac{\Rcal_{y,k}(\ell \circ \hFcal)}{n}+M \sqrt{\frac{\ln(|\Ycal||\Kcal|/\delta)}{2n}}\sum_{y \in \Ycal}\sum_{k\in I_{\Kcal}^y} \sqrt{\frac{|\Ical_{k}^{y}|}{n}}
\end{align*}
where we used the fact that $\sum_{y \in \Ycal}\sum_{k\in I_{\Kcal}^y}|\Ical_{k}^{y}|=n$. Moreover, we have that using the Cauchy--Schwarz inequality,  

$$
 \sum_{y \in \Ycal}\sum_{k\in I_{\Kcal}^y} \sqrt{\frac{|\Ical_{k}^{y}|}{n}} \le  \sqrt{\sum_{y \in \Ycal}\sum_{k\in I_{\Kcal}^y}\frac{|\Ical_{k}^{y}|}{n}} \sqrt{\sum_{y \in \Ycal}\sum_{k\in I_{\Kcal}^y} 1} \le\sqrt{|\Ycal||\Kcal|} .
$$
Thus, since $\ln({|\Ycal||\Kcal|}/\delta)\le \zeta\ln(e/\delta)$ with $\zeta =\max(1, \ln|\Ycal||\Kcal|)$ (see above), for any $\delta>0$, with probability at least $1-\delta$,
\begin{align}\label{eq:5}
&\frac{1}{n}\sum_{y \in \Ycal}\sum_{k\in I_{\Kcal}^y}|\Ical_{k}^{y}|\left(\EE_{z}[\ell(\hf^S(x),y_{})|z\in  \Ccal_{k}^{y}]-\frac{1}{|\Ical_{k}^{y}|}\sum_{i \in \Ical_{k}^y}\ell(\hf^S(x_{i}),y_{i}) \right) 
\\ \nonumber & \le2 \sum_{y \in \Ycal}\sum_{k\in I_{\Kcal}^y}|\Ical_{k}^{y}|\frac{\Rcal_{y,k}(\ell \circ \hFcal)}{n}+c \sqrt{\frac{\ln(e/\delta)}{2n}}.
\end{align}
For the  case of  $f=\hf^S$, by combining equation \eqref{eq:4} and \eqref{eq:5} with union bound, it holds that any $\delta>0$, with probability at least $1-\delta$, \begin{align*}
&\EE_{z,\epsilon}[\ell(\hf^S(g_\epsilon(x)),y_{})]  - \frac{1}{n} \sum_{i=1}^n \ell(\hf^S(x_{i}),y_{i})
\\ & \le  2 \sum_{y \in \Ycal}\sum_{k\in I_{\Kcal}^y}|\Ical_{k}^{y}|\frac{\Rcal_{y,k}(\ell \circ \hFcal)}{n}+c  \sqrt{\frac{\ln(2e/\delta)}{2n}}+c  \sqrt{\frac{2\ln(2e/\delta)}{n}} 
\\ & \quad + \frac{1}{n}\sum_{y,k}|\Ical_{k}^{y}|\EE_{x\sim \Dcal_{x|y},\epsilon}\left[\ell(\hf^S(g_\epsilon(x)),y_{})-\ell(\hf^S(x),y_{})|x\in  \Ccal_{k} ^{}\right].
\end{align*} 
We now consider the   case of $f=f^S_\theta$.  For the first term in the right-hand side of equation \eqref{eq:4}, 
\begin{align} \label{eq:6}
&\EE_{z}[\ell(f^S_\theta(x),y_{})|z\in  \Ccal_{k}^{y}]-\frac{1}{|\Ical_{k}^{y}|}\sum_{i \in \Ical_{k}^y}\ell(f^S_\theta(x_{i}),y_{i})
\\ \nonumber & =\EE_{z}[\ell( (d_\delta \circ v_\gamma \circ k_\beta \circ z_\alpha)(x),y_{})|z\in  \Ccal_{k}^{y}]-\frac{1}{|\Ical_{k}^{y}|}\sum_{i \in \Ical_{k}^y}\ell( (d_\delta \circ v_\gamma \circ k_\beta \circ z_\alpha)(x_{i}),y_{i})
 \\ \nonumber & =\ell( (d_\delta \circ v_\gamma )(\Kcal_k),y_{})-\ell( (d_\delta \circ v_\gamma )(\Kcal_k),y_{})=0.  
\end{align}
For the  case  of $f=f^S_\theta$, by combining equation \eqref{eq:4} and \eqref{eq:6}, it holds that any $\delta>0$, with probability at least $1-\delta$, \begin{align*}
&\EE_{z,\epsilon}[\ell(f^S_\theta(g_\epsilon(x)),y_{})]  - \frac{1}{n} \sum_{i=1}^n \ell(f^S_\theta(x_{i}),y_{i})
\\ & \le  c  \sqrt{\frac{2\ln(2e/\delta)}{n}} + \frac{1}{n}\sum_{y,k}|\Ical_{k}^{y}|\EE_{ x\sim \Dcal_{x|y},\epsilon}\left[\ell(f^S_\theta(g_\epsilon(x)),y_{})-\ell(f^S_\theta(x),y)|x\in  \Ccal_{k} ^{}\right]
\end{align*} 
Moreover, for the  case  of $f=f^S_\theta$, if  $x\in \Ccal_k \Rightarrow g_\epsilon(x) \in \Ccal_k$ with probability one, then   
\begin{align*}
\EE_{ x\sim \Dcal_{x|y},\epsilon}\left[\ell(f^S_\theta(g_\epsilon(x)),y_{})-\ell(f^S_\theta(x),y)|x\in  \Ccal_{k} ^{}\right]
 & =\ell( (d_\delta \circ v_\gamma )(\Kcal_k),y_{})-\ell( (d_\delta \circ v_\gamma )(\Kcal_k),y)
 \\ & =0.
\end{align*}
\end{proof}

\section{Why Do Random Down-Projection Matrices Help?} 
\label{app:random-matrix}
In addition to the flexibility in reducing the term $\frac{1}{n} \sum_{i=1}^n \ell(f(x_{i}),y_{i})$ when having multiple codebooks, the benefit from random projections in terms of \Cref{thm:1} is that this projection can reduce the term of $\Bcal_{g_{\epsilon}}(f)$. For example, consider $z + v$ where $z$ is an embedding and $v$ is some perturbation of $z$. When computing a projection with a random matrix $R$ by $R (z + v)$ and $v$ is in the null space of $R$,  then $R (z + v) = R z$ and hence this perturbation has no effect. Since $R$ is for down-sampling, there is a large null space for $R$. In other words, if the distribution shift results in $v$ such that $R v$ is small, $\Bcal_{g_{\epsilon}}(f)$ is small.

\section{Implementation Details.}
\label{sec:implementation_details}

Below we summarize all relevant implementation details.\footnote{\texttt{https://github.com/ftraeuble/experiments\_discrete\_key\_value\_bottleneck}.}

\paragraph{Model Architecture.} Our experiments build upon the following model architecture. \textbf{Encoder:} We use the following publicly available backbones:
\begin{enumerate}
    \item ResNet50 backbone, pre-trained supervised on ImageNet \citep{he2016deep, russakovsky2015imagenet}: We used the weights from PyTorchs  \verb|torchvision.models| package, specifically the weights under identifier \verb|ResNet50_Weights.IMAGENET1K_V2|
    \footnote{\texttt{https://pytorch.org/vision/stable/models.html}}. The backbone yields an embedding dimension of $m_z=2048$.
    \item ViT-B/32 backbone, CLIP-pre-trained \citep{radford2021learning}: We used the ViT B/32 model that can be easily obtained via OpenAI's official github repository \verb|clip| python package \footnote{\texttt{https://github.com/openai/CLIP}}. The backbone yields an embedding dimension of $m_z=512$.
    \item ResNet50w2 with twice the standard wideness, SwAV self-supervised pre-trained \citep{caron2020unsupervised}: We use the official pytorch model that can be loaded with PyTorch using the command \verb|torch.hub.load('facebookresearch/swav:main', 'resnet50w2')|. The backbone yields an embedding dimension of $m_z=2048$.
    \item ResNet50, DINO self-supervised pre-trained \cite{caron2021emerging}:  We use the official pytorch model that can be loaded with PyTorch using the command \verb|torch.hub.load('facebookresearch/dino:main', 'dino_resnet50')|. The backbone yields an embedding dimension of $m_z=1024$.
    \item As a fifth model, we present results using a ConvMixer pre-trained on a downsampled (32x32) Imagenet training set. As we wanted to directly compare against SDMLPs performance on the same class-incremental learning setting, we use their ConvMixer embeddings evaluated on the entire Imagenet and CIFAR10 dataset, which can be downloaded from the associated github repository.\footnote{\texttt{https://github.com/anon8371/AnonPaper1}} The backbone yields an embedding dimension of $m_z=256$.
\end{enumerate}

To cover a broad range of feature embeddings, we extracted the spatial continuous representation after the second last layer of residual blocks in the case of the two self-supervised pre-trained backbones and applied adaptive average pooling. The inferred continuous-valued representations cover a broad range of embedding dimensionalities depending on the backbone. In all our experiments we obtained the $C$ embedding heads for each key-value codebook by simply down-projecting this vector using $C$ random projection matrices into $C$ heads of dimension $d_{key}$. \textbf{Key-Value Bottleneck:} The key-value bottleneck consists of $256$ key-value codebooks that have each $4096$ key-value pairs per codebook. Keys are of the same dimension as the embedding heads ($d_{key}$), and we chose the value codes to be of the same size as the classes to predict, i.e.\, $d_{value} = 10$. \textbf{Decoder:} All $C$ fetched value codes are element-wise average-pooled at the end and the resulting tensor is passed to a softmax layer that predicts the class label.
For simplicity of exposition, we assume that the encoder representation does not have spatial dimensions; nevertheless, the extension to spatial feature maps is straightforward.

\paragraph{Initialization of Keys.} To initialize the keys, we build upon the exponential moving average updates (EMA) implemented in the pip package \verb|vector-quantize-pytorch|\footnote{\texttt{https://github.com/lucidrains/vector-quantize-pytorch}} as introduced in \cite{razavi2019generating, van2017neural}. For a given sequence of $t=1, ..., T$ mini-batches, we compute the moving averages of the codes (=key) positions $k^{(t)}_i$ and counts $N^{(t)}_i$ as follows for the c-th codebook
\begin{align}
N^{(t)}_i &:= \gamma N^{(t-1)}_i + n^{(t)}_i (1 - \gamma), \\
m^{(t)}_i &:= \gamma m^{(t-1)}_i + \sum_j^{n^{(t)}_i} E^c(x)^{(t)}_{i,j} (1 - \gamma), \\
k^{(t)}_i &:= \frac{m^{(t)}_i}{N^{(t)}_i}
\end{align}
where $E^c(x)^{(t)}_{i,j=1,...,n^{(t)}_i}$ are the $n^{(t)}_i$ (head) embeddings of the set of observations in the mini-batch that attach to the i-th key. We use a decay factor of $\gamma=0.95$. Moreover, we expire codes if they are not being adequately utilized. If the cluster size corresponding to a key (i.e., the moving average of the number of encodings that attach to a key) is below a threshold, we reinitialize the said key to a randomly selected encoding. This threshold scales as: $0.1 \cdot \text{batch-size} \cdot h \cdot w / \text{num-pairs}$
with $h$ and $w$ the spatial feature map dimensions in case of spatial pooling after the bottleneck. Keys are initialized at random encoder feature positions from the first batch. Our experiments apply spatial pooling before the bottleneck such that $h=w=1$.

\paragraph{Training.} After initializing the keys for 10 epochs on the key-initialization dataset (1 epoch on the Imagenet datasets), we train the model on the class-incremental CIFAR10 task with the SGD PyTorch optimizer without any weight decay or momentum and a learning rate of $lr = 0.3$ for the bottleneck and $lr = 0.001$ for the linear probe. We used a label smoothing parameter of 0.1 though we did not find its value to be sensitive to the result. We use no learning rate schedule. We used a batch size of 256 during key initialization and continual learning.\footnote{To the best of our knowledge, all software and assets we use are open source and under MIT, Apache or Creative Commons Licenses.}

\section{Additional Experiments}
\label{app:additional-experiments}

\subsection{Granularity During Class-Incremental Learning.} 
\begin{figure}
    \centering
    \includegraphics[width=0.5\textwidth]{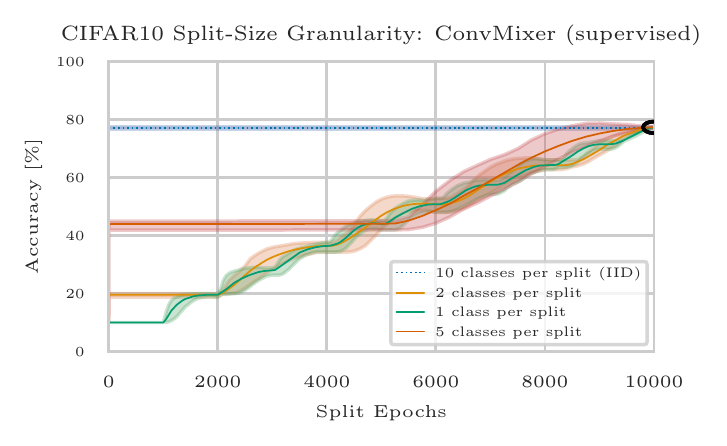}
    \caption{Analysing the effect of class-granularity during class incremental learning. In line with our theory, we observe \textit{across all possible class-split sizes} the appealing learning property that the bottleneck yields the same performance at the end of training.
    \looseness=-1}
    \label{fig:fig_granularity}
\end{figure}

In addition to the class-incremental learning scenario discussed in the main text, we experimented with additional class-incremental curricula of different split granularities. Specifically, we also tested the same model's performance (with the ConvMixer backbone) not only on 5 phases with 2 classes each but also on the more extreme case of 10 phases with a single class each and two phases with 5 classes each. We show these results in \Cref{fig:fig_granularity}, further supporting our finding from the main experiment that the proposed model is robust against various class-incremental curricula and approaches the same performance at the end of training.

\subsection{Additional Experiments on ConvMixer Backbone.} 
\begin{figure}
    \centering
    \includegraphics[width=0.5\textwidth]{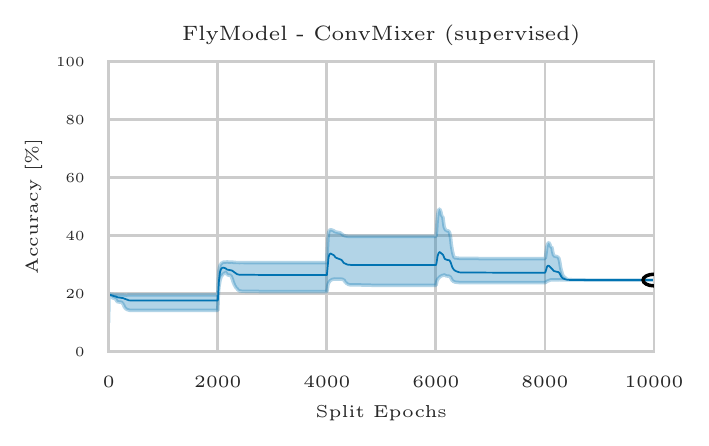}
    \caption{Applying the FlyModel from \citep{shen2021algorithmic} to the CIFAR10 learning setup with each split being repeated 2000 epochs. The model with hyperparameters from \citep{sdm_paper} learns new classes at the beginning of a new split but quickly drops to some lower-level performance until the end of the phase. Eventually, the model cannot integrate any new information and remains at a very low level.
    \looseness=-1}
    \label{fig:fly_model}
\end{figure}

The authors in \citep{sdm_paper} report two further methods with higher performance but their "SDMLP+EWC" results does not apply as it requires task boundaries and the FlyModel from \citet{shen2021algorithmic} was not trained on the batched 2000 epochs streaming setup. However, we also attempted to implement the FlyModel that proposes a solution for continual learning inspired by the neural circuit in the fruit fly olfactory system.  Similar to us, the FlyModel also relies on a frozen backbone but then projects this embedding via sparse binary projections (which are likewise fixed) into an extremely high-dimensional space reflecting more than 10,000 so-called "Kenyon cells". It then instantiates num-class "MBON cells" that are associative layers, fed from this high-dimensional representation. The FlyModel is trained such that learning a particular class updates the "MBON" associative layer of this class and was primarily tested under less challenging task-incremental settings by the authors. As mentioned before, the model reported in \citep{sdm_paper} was not trained for 2000 epochs per split. Instead, it was only updated over a single epoch simply assuming no further catastrophic forgetting in the batched streaming setting we are investigating (the FlyModel is not trained using back-propagation). In order to make the comparison as fair as possible, we trained the FlyModel for 2000 epochs per split building upon the implementation in \cite{sdm_paper}. Following \citet{sdm_paper}, we applied the following parameters that were reported to work best for them on this task and this backbone, specifically a learning rate of 0.005, 10,000 Kenyon cells and the number of projection neuron connections to be 32. We also set $\alpha=0$ to minimize catastrophic forgetting as described by \citet{shen2021algorithmic}. After rerunning this model for 5 random seeds each we obtained the results presented in \Cref{fig:fly_model}. As we can see, the model learns new classes at the beginning of a new split but quickly drops to some lower-level performance until the end of the phase. Eventually, the model cannot integrate any new information and remains at a very low level, which indicates strong catastrophic forgetting in the batched streaming setting.

\subsection{Additional Baseline: Ridge Regression on Frozen Backbone Features}
\label{app:ridge_regression}
\begin{figure}
    \centering
    \includegraphics[width=1.0\textwidth]{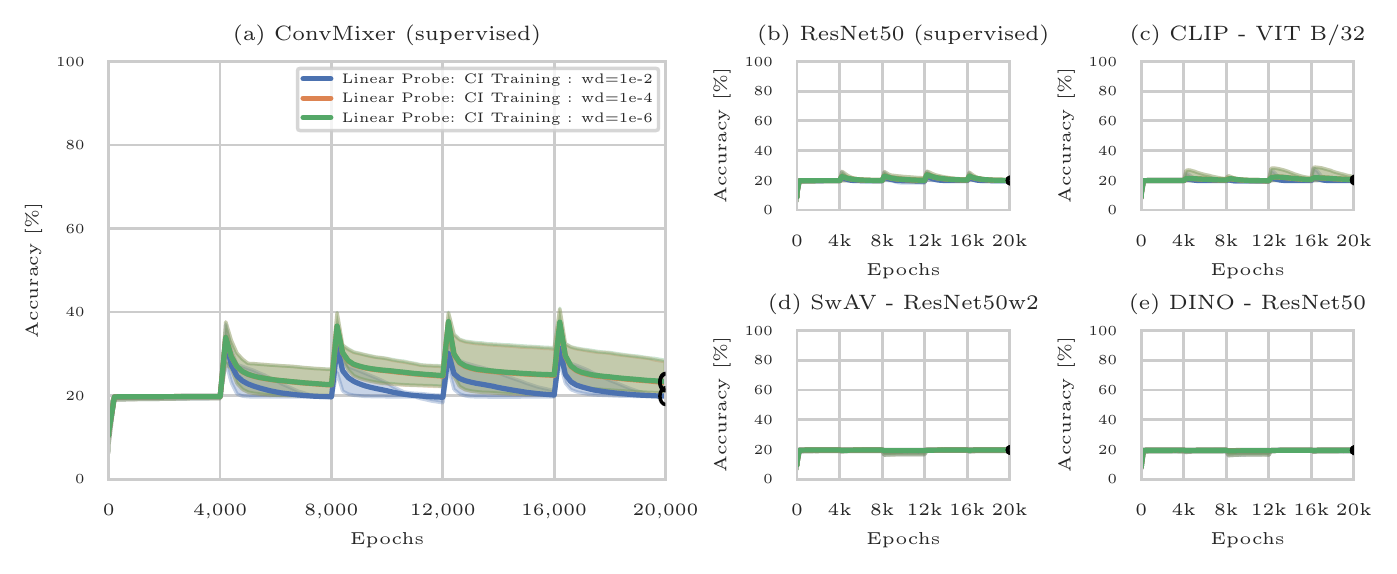}
    \caption{Additional baseline experiments: Rerunning the CIFAR10 experiment on all five backbones using the same Linear Probe as in Figure 3. In contrast to these baselines already shown in the main paper, we now applied weight decay on the head parameters across four orders of magnitude with $w_d \in [10^{-2}, 10^{-4}, 10^{-6}]$. The learned models are unable to effectively integrate novel class information using this additional form of ridge regularization.
    \looseness=-1}
    \label{fig:fig_ridge_lp}
\end{figure}

\begin{figure}
    \centering
    \includegraphics[width=1.0\textwidth]{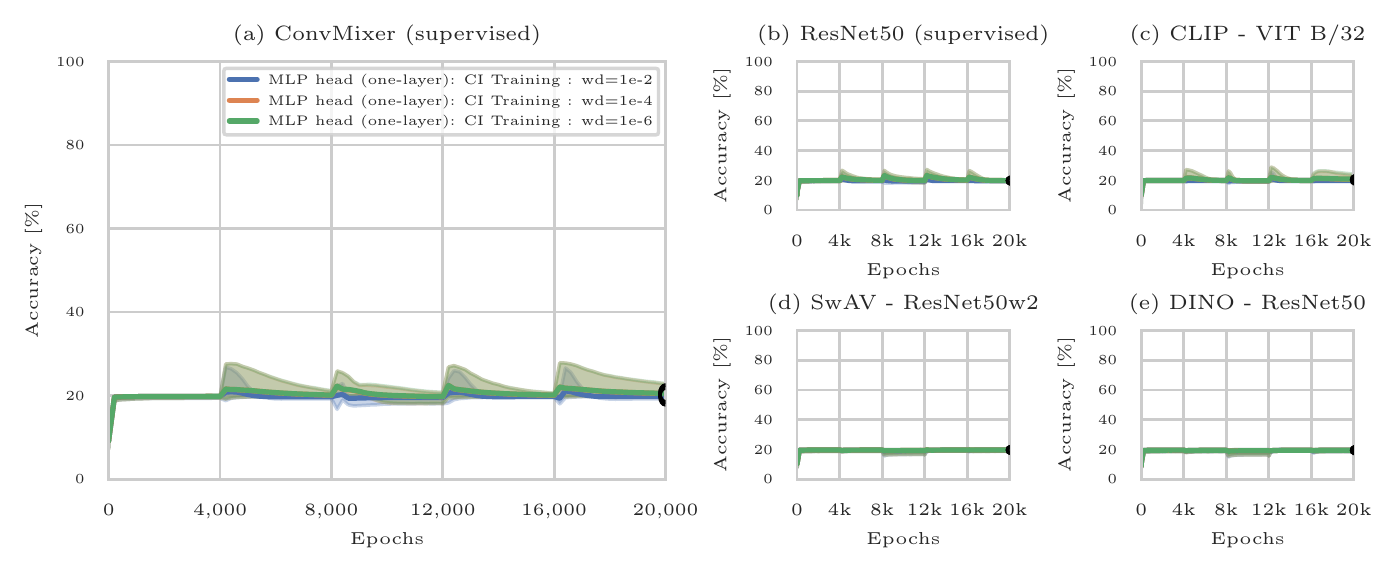}
    \caption{Additional baseline experiments: Rerunning the CIFAR10 experiment on all five backbones using the same MLP head as in Figure 3. In contrast to these baselines already shown in the main paper, we now applied weight decay on the head parameters across four orders of magnitude with $w_d \in [10^{-2}, 10^{-4}, 10^{-6}]$. The learned models are unable to effectively integrate novel class information using this additional form of ridge regularization.
    \looseness=-1}
    \label{fig:fig_ridge_mlp}
\end{figure}

We have also tested a further baseline method by directly performing ridge regression with the frozen feature encoder. Specifically, we were rerunning the CIFAR10 experiment on all five backbones using two different heads (Linear Probe and as well as the MLP head from Figure 3). In contrast to these baselines already shown in the main paper, we now applied weight decay on the head parameters across four orders of magnitude with $w_d \in [1e-2, 1e-4, 1e-6]$. We repeated three runs with different random seeds each. In total, we trained 90 additional models for this experiment. As illustrated in \Cref{fig:fig_ridge_lp} in the case of the Linear Probe and \Cref{fig:fig_ridge_mlp} in the case of the MLP head, the learned networks were unable to effectively integrate novel class information using this additional form of ridge regularization.

\subsection{Class-Incremental Learning on Larger Datasets.}
\begin{figure}
    \centering
    \includegraphics[width=0.7\textwidth]{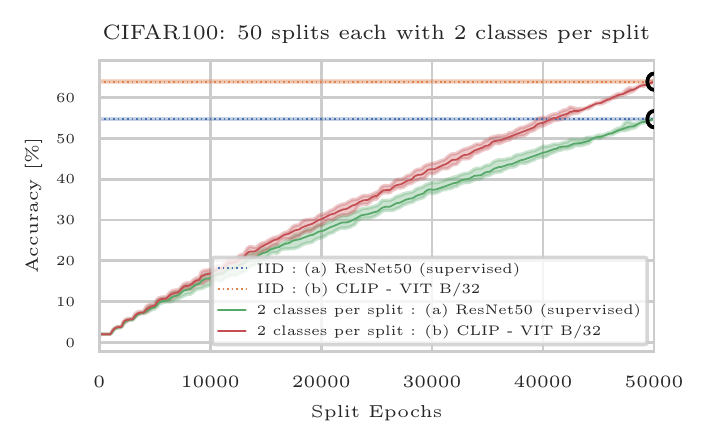}
    \caption{Additional class-incremental learning experiment on CIFAR100: Using 50 splits, each consisting of 2 classes, we tested our model on the ResNet-50 and CLIP-Backbone architectures. Our model achieved final accuracies of 54.7\% and 64\% respectively, compared to the 43\% achieved by SDMLP on the same task \citep{sdm_paper}. Keys were initialized on the CIFAR10 dataset for this experiment.
    \looseness=-1}
    \label{fig:fig_cifar100}
\end{figure}

\begin{figure}
    \centering
    \includegraphics[width=0.7\textwidth]{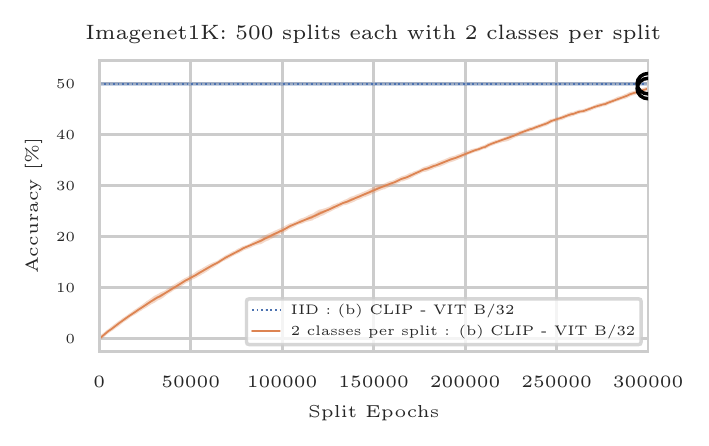}
    \caption{Additional class-incremental learning experiment on ImageNet: We evaluated our model on 500 splits, each comprising 2 classes, using the CLIP encoder. The final accuracy reached 49.9\% for the iid setting and 49.0\% for the class-incremental setting. To the best of our knowledge, no other method has been tested on such a large-scale class-incremental learning setting, encompassing 1000 classes and 500 splits. Keys were initialized on the CIFAR100 dataset for this experiment.
    \looseness=-1}
    \label{fig:fig_imagenet}
\end{figure}

To further test the future scalability of our method, we have run additional experiments on two larger datasets with many more classes. Importantly, we were using the very same architecture as for the CIFAR10 experiments:

\begin{enumerate}
    \item CIFAR-100 (see \Cref{fig:fig_cifar100}): Using 50 splits, each consisting of 2 classes, we tested our model on the ResNet-50 and CLIP-Backbone architectures. Our model achieved final accuracies of 54.7\% and 64\% respectively, compared to the 43\% achieved by SDMLP on the same task \citep{sdm_paper}.
    \item ImageNet (see \Cref{fig:fig_imagenet}): We evaluated our model on 500 splits, each comprising 2 classes, using the CLIP encoder. The final accuracy reached 49.9\% for the iid setting and 49.0\% for the class-incremental setting. To the best of our knowledge, no other method has been tested on such a large-scale class-incremental learning setting, encompassing 1000 classes and 500 splits.

\end{enumerate}

\newpage

\section{Additional Analysis}
\label{app:additional-analysis}

\subsection{Robustness via Discrete Bottleneck.}
\begin{figure}
    \centering
    \includegraphics[width=1.0\textwidth]{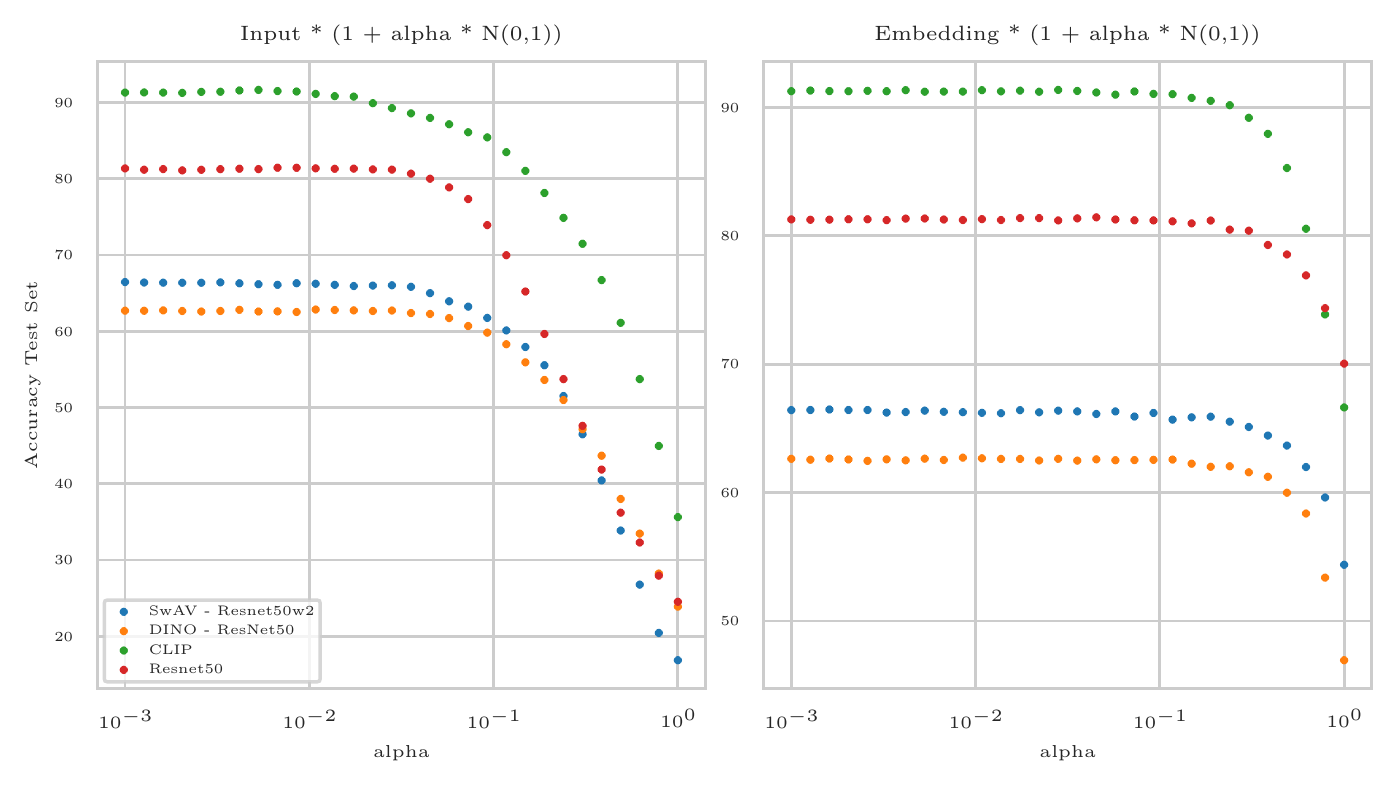}
    \caption{We have conducted additional experiments to assess the models' robustness. Specifically, we evaluated the performance of the models analyzed in Figures 8-11, which employ different backbones, when subjected to additive Gaussian noise spanning three orders of magnitude. We applied the noise either to the original input pixels (left) or the backbone embeddings (right). Results indicate that the model's performance remains largely unaffected by the noise and only starts to deteriorate substantially when the noise level exceeds $\alpha=0.1$, at which point the image becomes nearly unrecognizable upon visual inspection.
From a theoretical per
    \looseness=-1}
    \label{fig:fig_robustness}
\end{figure}

The discretization of the bottleneck confers additional robustness, which is a significant attribute of our model. To illustrate this property, we have conducted additional experiments to assess the models' robustness. Specifically, we evaluated the performance of the models analyzed in \Cref{fig:fig_ana_1,fig:fig_ana_2,fig:fig_ana_3,fig:fig_ana_4,fig:fig_ana_5}, which employ different backbones, when subjected to additive Gaussian noise spanning three orders of magnitude. We applied the noise either to the original input pixels or the backbone embeddings. Our results which are shown in \Cref{fig:fig_robustness}, indicate that the model's performance remains largely unaffected by the noise and only starts to deteriorate substantially when the noise level exceeds $\alpha=0.1$, at which point the image becomes nearly unrecognizable upon visual inspection.

\subsection{Value Codes}

Next, we want to provide additional analysis on the similarity of the learned values when training iid versus class-incrementally. To answer how much the value codes differ depending on the training order (everything else being the same), we computed the relative mean absolute difference between values learned iid and values learned class-incrementally in the case of the five analyzed models in the previous section, formally
\begin{align}
    \frac{\sum|values_{CI} - values_{IID}|}{\sum|values_{CI}|}.
\end{align}
As we can see from \Cref{tab:value_similarity}, the analysis suggests that the learned values are indeed fairly similar, albeit not identical, explaining the almost identical final performance.

\begin{table}[]
    \centering
    \begin{tabular}{l|c}
    \toprule
    Backbone & Rel. mean abs. distance \\
    \midrule
    Resnet50 & 2.19\% \\
    CLIP & 2.09\% \\
    DINO & 2.2\% \\
    SwAV & 2.2\% \\
    ConvMixer & 3.07\% \\
    \bottomrule
    \end{tabular}
    \caption{Relative mean absolute distance between values trained iid and class-incremental for the five different backbones from the base experiment.}
    \label{tab:value_similarity}
\end{table}

\subsection{Key Codes}
\label{app:key-analysis}
In this section, we analyze key codes used in the proposed model to better understand how they cover the embedding manifold of the data. We investigate three questions: 
\begin{enumerate}
\item Do key codes broadly cover the embedding manifold with respect to the key initialization dataset?
\item How does key initialization on another dataset affect coverage on the embedding manifold under the target dataset?
\item Can we quantify the amount of key-value pair sharing among training samples?
\end{enumerate}

\begin{figure*}
    \centering
    \includegraphics[width=0.8\textwidth]{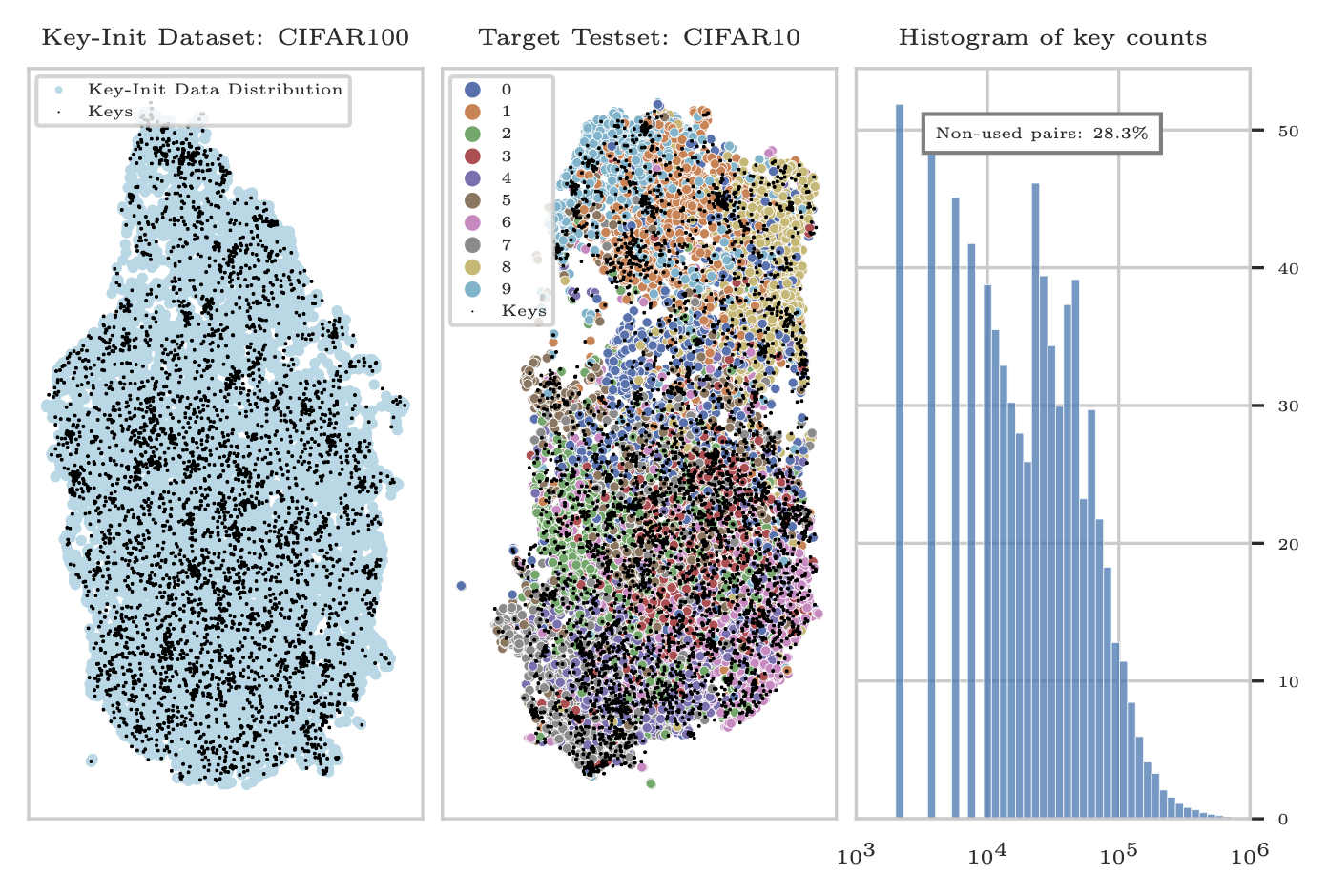}
    \caption{Analysing the \textbf{key codes} on the \textbf{ResNet50 backbone.} Left Plot: UMAP of the codebook embedding heads from the key initialization dataset (light blue) with the discrete key codes (black dots) as obtained by EMA. Keys are broadly covering the data manifold. Middle Plot. The key code distribution on the target test dataset. Keys exhibit increased clustering, which reflects the smaller feature diversity present in CIFAR10. Right Plot: Key code utilization across all codebooks over the course of class-incremental training (2000 epochs). The majority of keys is being shared among many different input samples, while the bottleneck retains some unused capacity.
    \looseness=-1}
    \label{fig:fig_ana_1}
\end{figure*}

\begin{figure*}
    \centering
    \includegraphics[width=0.7\textwidth]{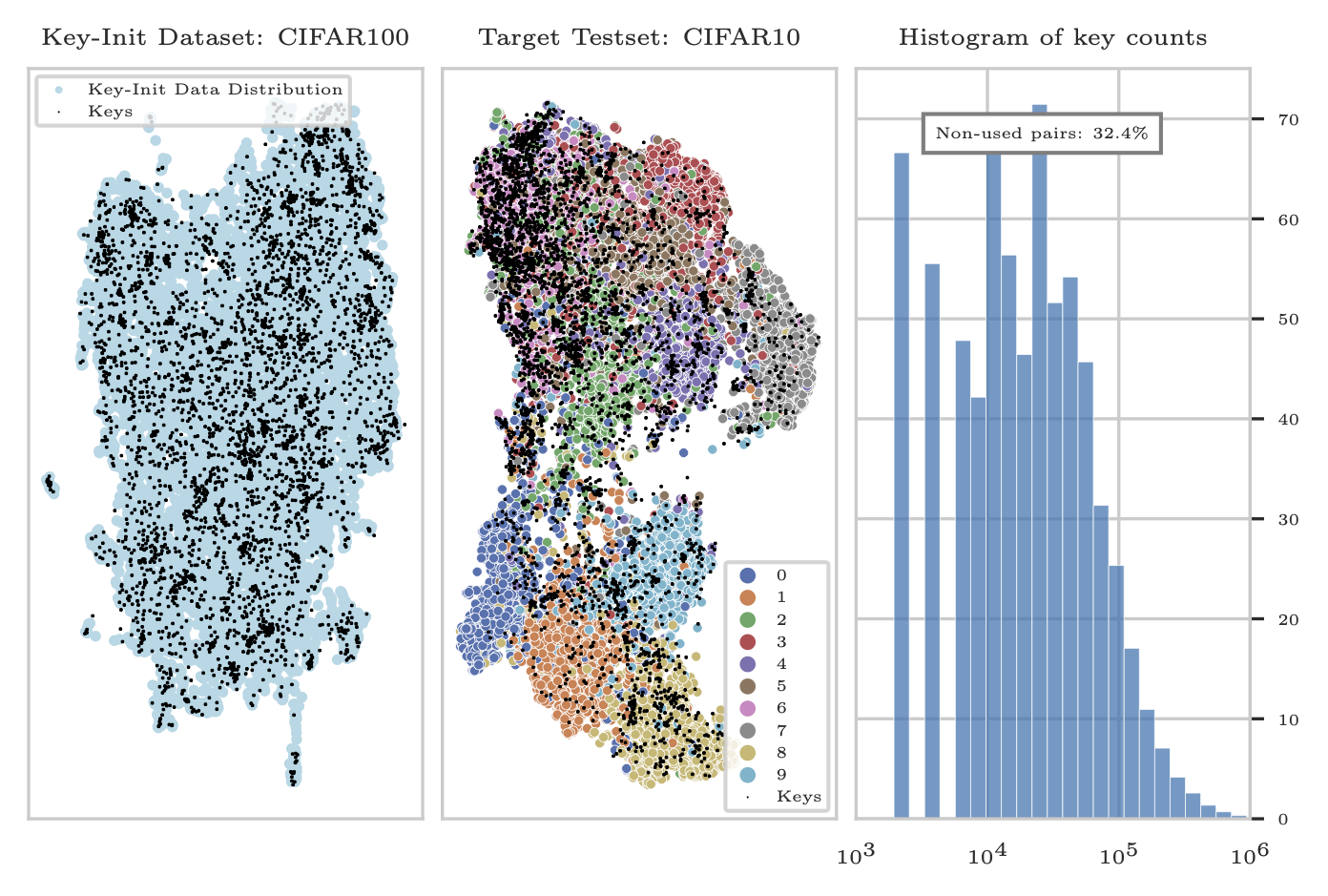}
    \caption{
    Analysing the \textbf{key codes} on the \textbf{CLIP-pretrained ViT backbone}. Left: UMAP of the codebook embedding heads from the key initialization dataset (light blue) with the discrete key codes (black dots) as obtained by EMA. Keys are broadly covering the data manifold. Middle. The key code distribution on the target test dataset. Although the coverage appears more clustered, the keys still cover the majority of the target test set, which reflects the smaller feature diversity present in CIFAR10. Right: Key code utilization across all codebooks over the course of class-incremental training (2000 epochs). The majority of keys is being shared among many different input samples, while the bottleneck retains some unused capacity.
    \looseness=-1}
    \label{fig:fig_ana_2}
\end{figure*}

\begin{figure*}
    \centering
    \includegraphics[width=0.7\textwidth]{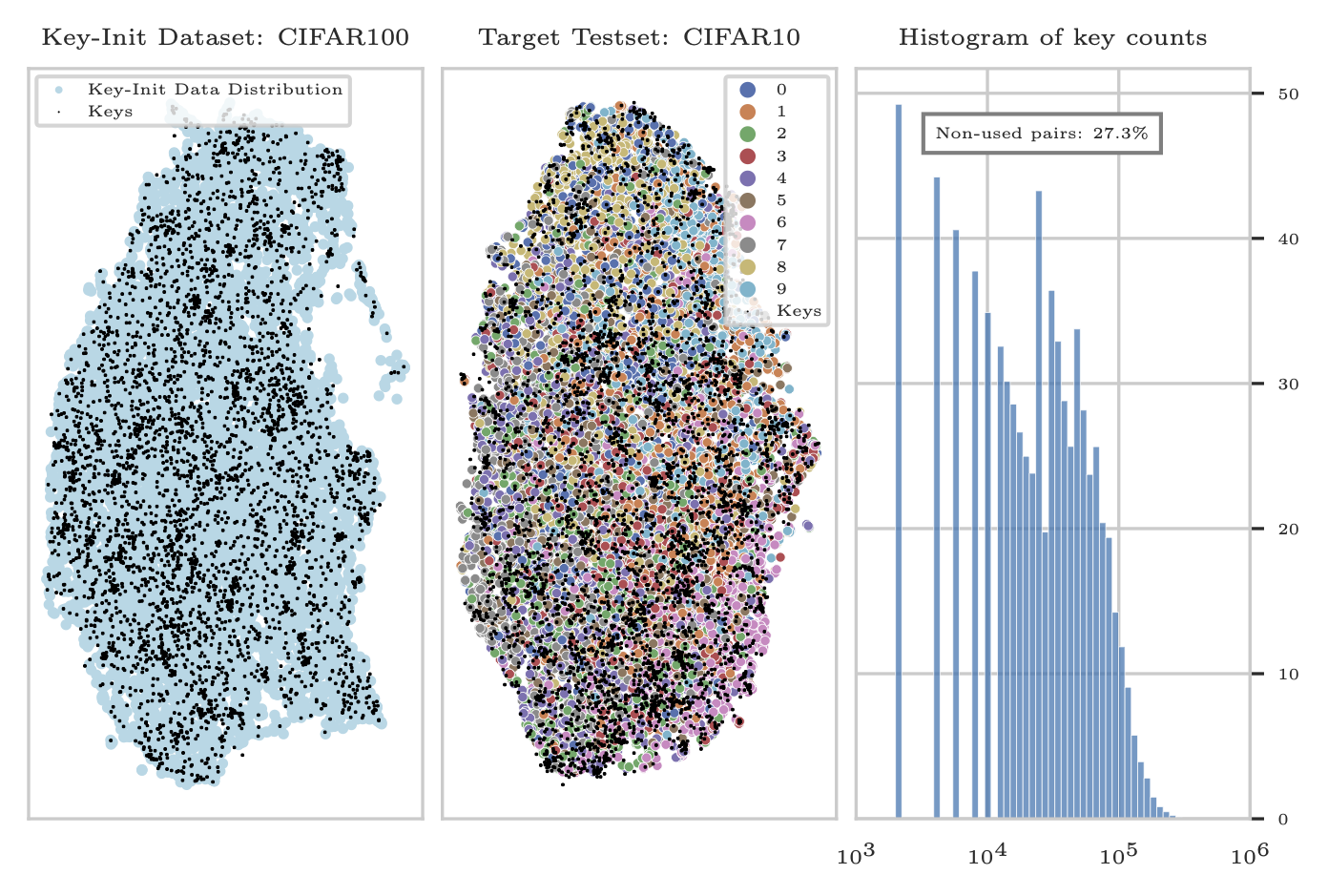}
    \caption{
    Analysing the \textbf{key codes} on the \textbf{DINO-pretrained backbone}. Left: UMAP of the codebook embedding heads from the key initialization dataset (light blue) with the discrete key codes (black dots) as obtained by EMA. Keys are broadly covering the data manifold. Middle. The key code distribution on the target test dataset. Although the coverage appears more clustered, the keys still cover the majority of the target test set, which reflects the smaller feature diversity present in CIFAR10. Right: Key code utilization across all codebooks over the course of class-incremental training (2000 epochs). The majority of keys is being shared among many different input samples, while the bottleneck retains some unused capacity.
    \looseness=-1}
    \label{fig:fig_ana_3}
\end{figure*}

\begin{figure*}
    \centering
    \includegraphics[width=0.7\textwidth]{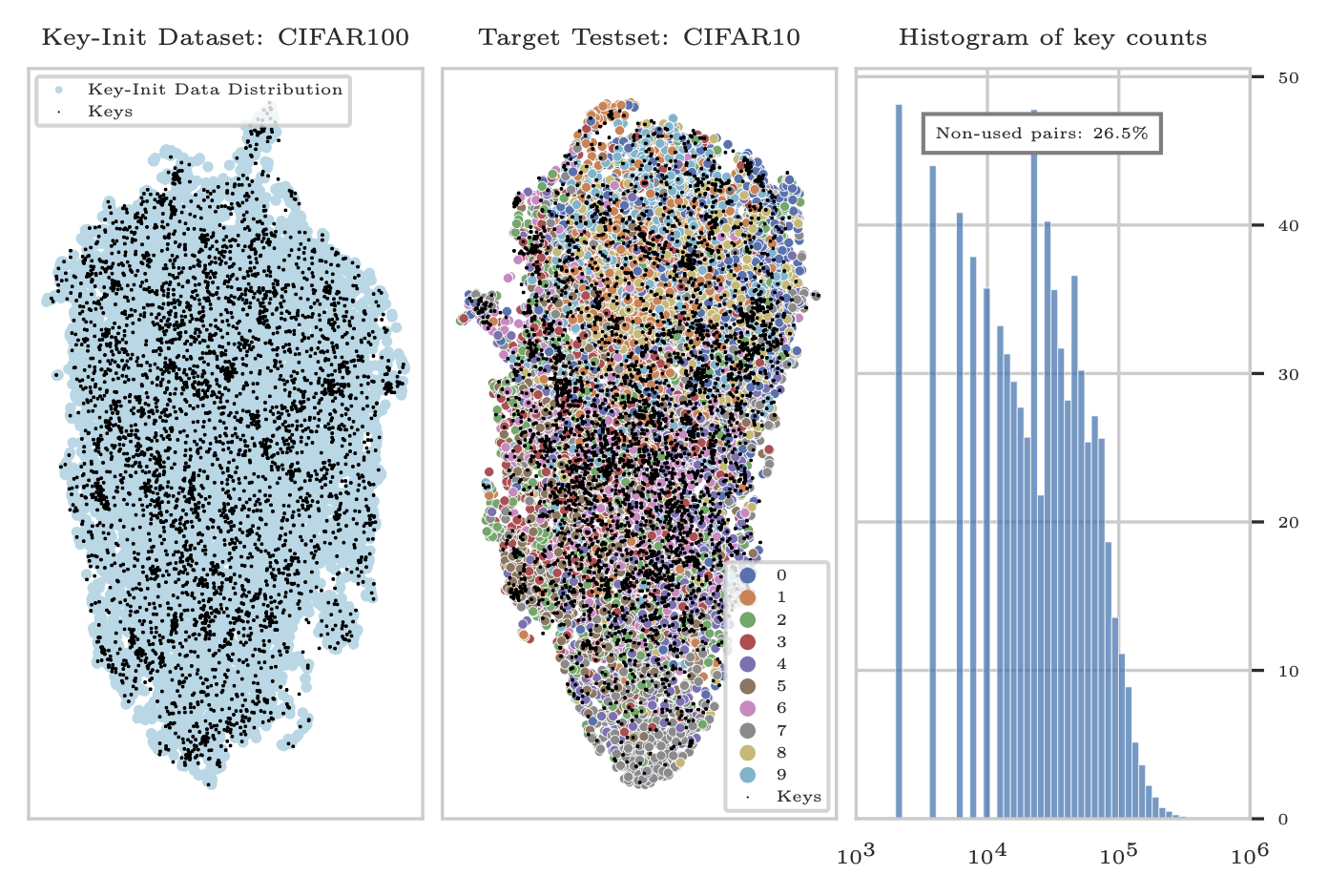}
    \caption{
    Analysing the \textbf{key codes} on the \textbf{SwAV-pretrained backbone.} Left: UMAP of the codebook embedding heads from the key initialization dataset (light blue) with the discrete key codes (black dots) as obtained by EMA. Keys are broadly covering the data manifold. Middle. The key code distribution on the target test dataset. Keys exhibit increased clustering, which reflects the smaller feature diversity present in CIFAR10. Right: Key code utilization across all codebooks over the course of class-incremental training (2000 epochs). The majority of keys is being shared among many input samples, while the bottleneck retains some unused capacity.
    \looseness=-1}
    \label{fig:fig_ana_4}
\end{figure*}

\begin{figure*}
    \centering
    \includegraphics[width=0.7\textwidth]{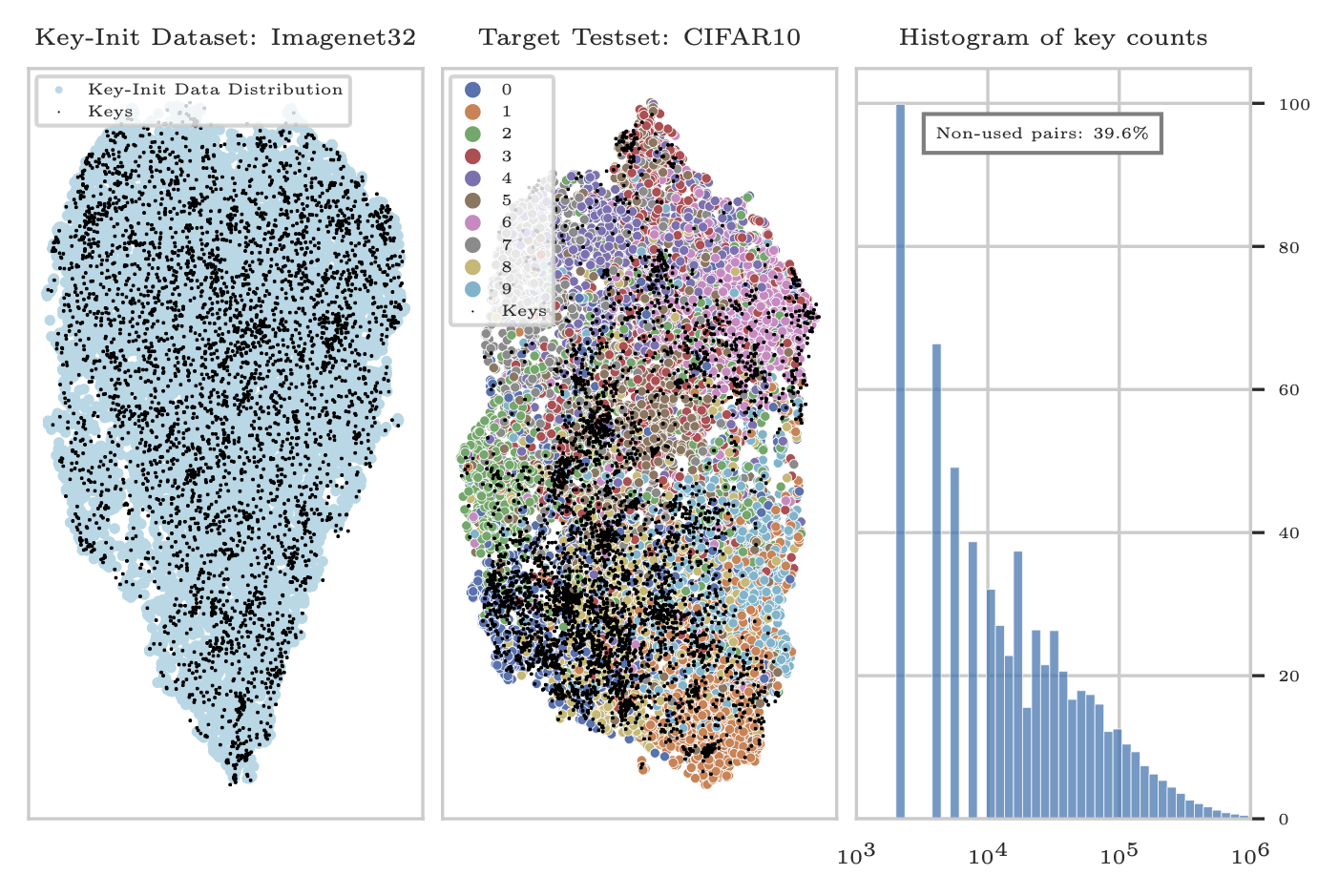}
    \caption{Analysing the \textbf{key codes} on the \textbf{ConvMixer backbone} from \citep{sdm_paper}. Left: UMAP of the codebook embedding heads from the key initialization dataset (light blue) with the discrete key codes (black dots) as obtained by EMA. Keys are broadly covering the data manifold. Middle. The key code distribution on the target test dataset. Keys exhibit increased clustering, which reflects the smaller feature diversity present in CIFAR10. Right: Key code utilization across all codebooks over the course of class-incremental training (2000 epochs). The majority of keys is being shared among many input samples, while the bottleneck retains some unused capacity.
    \looseness=-1}
    \label{fig:fig_ana_5}
\end{figure*}

To answer these questions, we inspect the key codes for each backbone discussed in \cref{sec:experiments}, analyzing the model per backbone with the same fixed random seed trained under the class-incremental training setting. Results are shown in \Cref{fig:fig_ana_1,fig:fig_ana_2,fig:fig_ana_3,fig:fig_ana_4,fig:fig_ana_5}. 

We begin by inspecting the key code space of the first codebook under the supervised ResNet50 backbone in \Cref{fig:fig_ana_1} (left plot). The keys of this model were initialized on the unlabelled CIFAR100 dataset. When applying the UMAP dimensionality reduction on the input heads of 10k CIFAR100 training samples, we can see that the 4096 key codes of this codebook (black dots) broadly cover the entire data manifold (light blue dots). This verifies that the EMA method is working as expected. When repeating the same procedure for the test set of the target distribution, here CIFAR10, we observe that the fixed keys from before are still covering the entire data domain of CIFAR10, see \Cref{fig:fig_ana_1} (middle plot). Compared to the distribution on the key initialization dataset, the coverage is more sparse with several concentrations of multiple clusters of keys. We hypothesize that these keys might have specialized on certain features present in CIFAR100 but not present in CIFAR10. At the same time, we observe keys across all CIFAR10 class domains (indicated by color) even though the same class labels are not present in CIFAR100. We observe this behaviour across all backbones: Interestingly, we qualitatively observe a slightly reduced clustering in the case of the lower-level SSL feature embeddings in \Cref{fig:fig_ana_3} and \Cref{fig:fig_ana_4} over the others. In our ablation experiments, we only observed a slight drop in performance when initializing the keys on the much higher-resolution ImageNet dataset, which can be partially explained with the analysis under the ConvMixer model, that was initialized on a 32x32 downsampled Imagenet variant (see \Cref{fig:fig_ana_5}). First, the EMA approach still yields a broad coverage of the much more diverse Imagenet data manifold without any change in EMA parameters. Second, the keys are covering the CIFAR10 domain sparsely but there are still at least a few keys present close to each domain. From our discussed results we can thereby conclude that the bottleneck can still effectively learn to generalize along all classes even though the keys might not be optimally distributed. Finally, we tracked the usage of all key-value pairs across all codebooks over the course of the training class incrementally on the target dataset; see \Cref{fig:fig_ana_1} (right plot). Note that the y-axis is scaling in counts of thousands. In the case of the ResNet50 backbone, we observe that the majority of key-value pairs were fetched between 10,000 and 150,000 times over the course of training. As we repeat each episode for 2000 epochs, each key that is snapped to by a particular training image contributes a utilization count of 2000 to this key. Interestingly, there is also a long tail of a few keys that are used very often and about 28\% of keys that are never used, which suggests that the bottleneck did not distil the information during train time in all available key-value codes. This behaviour can be seen similarly among all other backbones, too.

\end{document}